\documentclass[times, review, 10pt]{elsarticle}

\usepackage{amssymb}
\usepackage{amsmath}

\usepackage{url}
\usepackage[breaklinks=true]{hyperref} 
\usepackage{graphicx}
\usepackage{float}
\usepackage{tabularx}
\usepackage{subcaption} 
\usepackage{booktabs}
\usepackage{multirow}
\usepackage{caption}
\usepackage{makecell}

\journal{Pattern Recognition}

\begin{document}

\begin{frontmatter}

\title{Scalable Class-Incremental Learning Based on Parametric Neural Collapse}

\author{ Chuangxin Zhang, Guangfeng Lin,Enhui Zhao, Kaiyang Liao, Yajun Chen} 

 \address {
  Xi'an University of Technology ,
  5 South Jinhua Road, Xi'an ,710048,
  Shaanxi Province,
  China
}

\begin{abstract}
 Incremental learning often encounter challenges such as overfitting to new data and
 catastrophic forgetting of old data. Existing methods can effectively extend the model
 for new tasks while freezing the parameters of the old model, but ignore the necessity of
 structural efficiency to lead to the feature difference between modules and the class misalignment due to evolving class distributions. To address these issues, we propose scalable class-incremental
 learning based on parametric neural collapse (SCL-PNC) that enables demand-driven,
 minimal-cost backbone expansion by adapt-layer and refines the static into a dynamic
 parametric Equiangular Tight Frame (ETF) framework according to incremental class.
 This method can efficiently handle the model expansion question with the increasing number of categories in real-world scenarios. Additionally, to counteract feature
 drift in serial expansion models, the parallel expansion framework is presented with a
 knowledge distillation algorithm to align features across expansion modules. Therefore, SCL-PNC can not only design a dynamic and extensible ETF classifier to address
 class misalignment due to evolving class distributions, but also ensure feature consistency by an adapt-layer
 with knowledge distillation between extended modules. By leveraging neural collapse, SCL-PNC induces the convergence of the incremental expansion model through
 a structured combination of the expandable backbone, adapt-layer, and
 the parametric ETF classifier. Experiments on standard benchmarks demonstrate the
 effectiveness and efficiency of our proposed method.
Our code is available at \url{https://github.com/zhangchuangxin71-cyber/dynamic_ETF2}.
\end{abstract}

\begin{graphicalabstract}
\begin{figure}[H]
	\centering
	\includegraphics[width=1\textwidth,height=0.5\textheight]{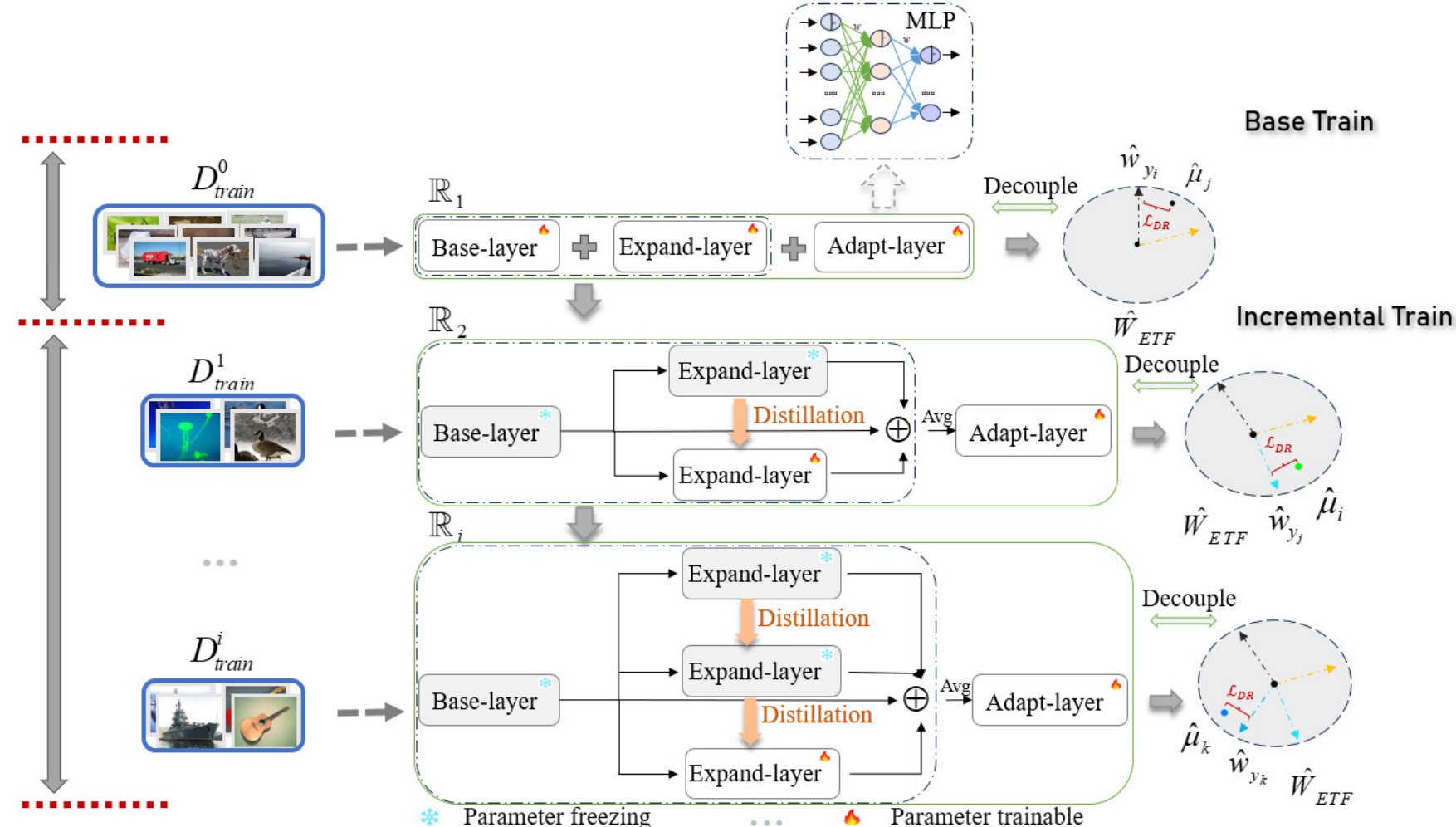}
	
\end{figure}
\end{graphicalabstract}

\begin{highlights}
\item An Adapt-Layer can constrain the feature vector prototypes of the model backbone outputs so that they align with the corresponding category classifier prototypes.
\item A Dynamic Parametric ETF classifier can address the misalignment due to evolving class distributions problem caused by the continuous increase in data categories.
\item Experiments demonstrate that SCL-PNC is superior to  the state-of-the-art methods on small-scale and large-scale datasets.
\end{highlights}


\end{frontmatter}



\section{Introduction}
\label{sec1}

Incremental Learning (IL) aims to continuously acquire new knowledge, similar to human learning, while simultaneously retaining and supplementing existing knowledge. IL is an effective method for addressing data stream changes (or non-stationary data distributions). Although many existing IL methods achieve continuous learning by expanding new model structures for each incremental task while freezing old parameters, they rarely assess whether structural expansion is truly necessary. This inevitably results in rapid growth of both parameter count and computational cost, and can sometimes exacerbate catastrophic forgetting \cite{kirkpatrick2017overcoming}. In practical applications, this limitation poses significant challenges, particularly in maintaining robust performance on old categories and mitigating feature drift between successive expansion modules.

Model expansion has emerged as a promising direction for incremental classification. Specifically, expanding backbone networks \cite{xu2018reinforced}\cite{yoon2017den}and fully connected layers \cite{yan2021der} can enhance classification and feature aggregation for new tasks. Other works focus on Knowledge Distillation (KD) \cite{zhang2020consolidation} to transfer knowledge from the old model to the new, primarily mitigating catastrophic forgetting. Furthermore, recent studies leverage CLIP \cite{li2021supervision} for incremental training, utilizing contrastive learning to achieve cross-modal alignment and improve model generalization \cite{wang2022sprompts}. The latest works focus on enhancing model adaptability and efficiency, such as \cite{9}, which explores dynamic feature space expansion, and CLSNet \cite{10}, which efficiently controls parameters without expanding the network size. These advancements demonstrate significant progress in handling non-stationary data distributions and mitigating catastrophic forgetting, thereby improving the practical value of incremental learning models.

\begin{figure}[ht]
	\centering
	\includegraphics[width=0.8\textwidth,height=0.5\textheight]{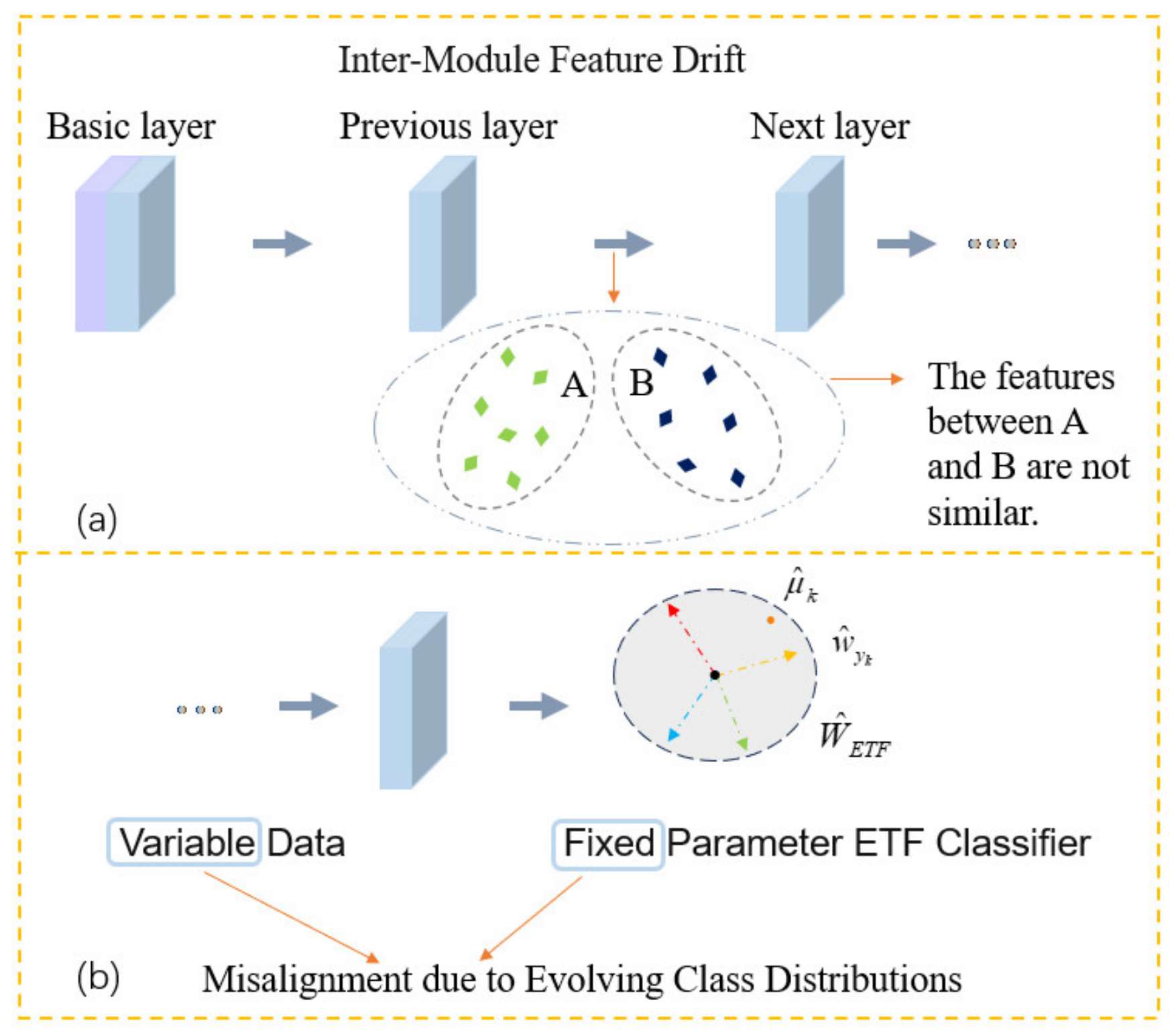}
	\caption{ The two issues of existing model-expansion methods. (a) illustrates the inter-module feature drift that arises during module alignment with a substantial discrepancy between category features A and B. Consequently, the knowledge transferred across modules becomes discontinuous ultimately leading to catastrophic forgetting of old classes. (b) shows the misalignment due to evolving class distributions. As the number of categories increases while the ETF classifier maintains fixed parameters, a pronounced distribution bias emerges.}\label{Fig.1}
\end{figure}

However, existing expansion-based methods mostly suffer from inter-module feature drift (or consistency degradation) across extended modules (as illustrated in Fig. \ref{Fig.1}(a). This issue significantly affects the robustness and classification accuracy of prior categories. Furthermore, due to the dynamic emergence of real-world data, existing static ETF classifiers require a fixed number of categories, leading to a misalignment due to evolving class distributions between the modules and the fixed parameter classifier (Fig. \ref{Fig.1}(b)). Therefore, it is crucial to develop a model that can dynamically adapt to new categories while simultaneously addressing both the inter-module feature drift and the misalignment due to evolving class distributions introduced by model expansion.

To mitigate feature drift during architectural expansion, we employ a knowledge distillation (KD) mechanism to transfer information across newly added modules and introduce a dynamic parametric ETF classifier to address the misalignment caused by evolving class distributions. However, sequential expansion frameworks still suffer from a progressive decline in representational similarity as more modules are appended, as shown in our CKA analysis (Fig.~\ref{fig7} indicates the higher CKA score with the better similarity). To address this limitation, we further propose a Parallel Knowledge Distillation (P-KD) architecture, which allows all extended modules to distill directly from the frozen base layer. This parallel design preserves high feature consistency across modules-achieving a CKA score of 0.85 compared with 0.52 under serial expansion—and substantially reduces inter-module feature drift throughout the expansion process.

This work introduces SCL-PNC, a framework designed to support incremental class expansion under dynamically growing architectures. Our approach addresses the coupled challenges of feature drift, prototype mismatch, and distribution instability. Therefore, the key contributions are:
\begin{itemize}
\item Adapt-Layer for feature alignment

We design a specialized Adapt-layer to align backbone features with classifier prototypes via prototype regression and knowledge distillation. Unlike existing calibration techniques (e.g., FCS \cite{li2024fcs}’s feature calibration network) that operate on a single backbone or passively compensate for drift \cite{yu2020semantic}, this layer is designed to mitigate inter-module feature drift induced by backbone expansion. It ensures consistent feature geometry across different incremental stages, addressing an important limitation of prior methods that do not explicitly combine drift mitigation with dynamic model expansion.

\item Dynamic Parametric ETF Classifier

Building on Neural Collapse theory \cite{papyan2020neuralcollapse}, we extend the conventional fixed-geometry ETF to a Dynamic Parametric form, where the classifier weight matrix is adaptively updated via simplex vertex projection as new classes are introduced. This mechanism effectively resolves the mismatch between fixed classifier geometries adopted by existing ETF-based methods and evolving class distributions. In contrast, expandable CIL approaches with static classifiers (e.g., DER \cite{yan2021der}, MEMO \cite{zhou2022memory}) lack mechanisms to adapt classifier geometry to incremental class growth, leading to performance degradation in long-term incremental learning.

\item  Lightweight parallel expansion

We propose a knowledge-distillation-augmented parallel expansion strategy that prioritizes feature geometry consistency. By utilizing a frozen base-layer as a shared geometric reference and enforcing inter-module knowledge distillation, newly added modules are constrained to a common feature space. Compared with sequential expansion methods (e.g., DER \cite{yan2021der}) that lack an explicit geometric anchor and suffer from accumulated feature drift, our design effectively suppresses representation drift while maintaining favorable parameter efficiency.

\item Strong empirical performance

Experiments on multiple benchmarks show that SCL-PNC consistently outperforms existing methods and highlights the advantages of parallel expansion over sequential alternatives.
\end{itemize}

\section{Related Work}
\label{sec2}

SCL-PNC is related to three core areas, which are class-incremental learning(CIL) based on dynamic models, the feature drift problem in continuous learning, and the theoretical method of the neural collapse phenomenon. These areas respectively correspond to the basic architecture, the module relationship, and the discriminative ability, which collectively construct the foundational framework of SCL-PNC.

\subsection{Class-incremental learning based on dynamic models}
\label{subsec1}

To achieve the model's ability to continuously learn new tasks, existing methods are mainly divided into two categories, which are the backbone network reusing and the local network extending. Methods based on backbone reuse typically retrain the entire network architecture to adapt the incremental data by extending the backbone network, adding gating mechanisms\cite{aljundi2017expert}, dynamically extending the model and aggregating features with a maximum fully connected layer\cite{rusu2016pnn}, and model compression via knowledge distillation\cite{foster2022feature}. The local network extending methods emphasise on the additional module connected with the backbone network to model the distribution of the incremental data by the construction of the memory buffer\cite{zhou2022memory} and the progressive expansion-compression framework\cite{schwarz2018progress}. The latest works tend to enhance the above ideas by multi-view CIL framework leveraging orthogonalization for robust adaptation \cite{28} and the incremental vocabulary learning using dynamic updates for domain-specific tasks\cite{29}.

Although the above works can accommodate the difference of the incremental data through task-specific model architecture expansion, they often neglect the consistency of feature representations across the growing model modules. This oversight induces significant feature drift, which subsequently exacerbates the catastrophic forgetting problem.

\subsection{Feature drift problem in continuous learning}
\label{subsec2}
The primary cause of feature drift is that the data distribution of the current task learned by the model occurs the serious deviation to the data distribution of the previous task learning. To tackle this issue, existing methods attribute to two categories depending on the explicit reconstruction of the class prototypes. One is the direct feature drift compensation and evaluation, and centers on feature stability enhancement to generate compensation signal\cite{yu2020semantic} or evaluate the degree of drift \cite{peng2024lightweight} by explicitly modeling or quantifying feature drift. Another is the implicit drift constraint via regularization or strategy optimization, and concentrates on constraining model update direction\cite{magistri2024elastic} or dynamically selecting relearning strategies\cite{campello2015acm} to indirectly alleviate drift effects and avoid explicit drift modeling. The latest works incline to model data dynamics via Koopman operators\cite{shi2024koopman} and a corticohippocampal circuits-based hybrid neural network (CH-HNN)\cite{shi2025hybrid}, which emulates these dual representations, significantly mitigating catastrophic forgetting in both task-incremental and class-incremental learning.

Although existing methods have made progress in multi-angle exploration, these methods can passively respond to drift and can not combine with dynamic model expansion strategies, which limits their scalability in long-term incremental scenarios.

\subsection{Neural collapse phenomenon for incremental learning}
\label{subsec3}
Neural collapse (NC)\cite{papyan2020neuralcollapse}is the phenomenon that the same-category features and different-category features of the last layer of classifiers collapse to their intra-class means and together form the vertex of the ETF classifier when the training loss is 0. Recent works have shown that neural collapse can be well applied to few-shot incremental learning\cite{chen2025fewshot}, zero-shot classification\cite{zhang2021learnware}, reinforcement learning\cite{zhang2020unbiased}, and continuous learning\cite{kirkpatrick2017overcoming}. To utilize the geometric conditions of ETF classifier for maximizing inter class space and minimizing intra class variance, SCL-PNC expect to induce neural collapse in the backbone of the scalable model to embed the neural collapse phenomenon into the extending model.

Because existing methods neglect the misalignment due to evolving class distributions between the sustainable growth data and the fixed parameter ETF classifier to lead to the knowledge forgetting problem between the expansion modules, SCL-PNC expands dynamic model strategy to effectively classify in the long-term growth data.

\section{Class-Incremental Learning}
\label{sec3}
Class-incremental learning methods aim to endow models with the capability of continuous learning, enabling them for adapting to evolve data streams. Suppose there exists a sequence of $T$ training tasks, which are $D^0$,$D^1$,..., and $D^{T}$ with non-overlapping classes, where $D^t=\{(x_i^t,y_i^t)|1\le i\le n_t\}$ represents the $t$-th training task. $x_i^t \in \mathbb{R}^d$ denotes the training sample, and ${y_i} \in {Y_t}$ stands for the corresponding class labels of $x_i^t$, and ${Y_t}$ is the label space of task $t$. ${Y_t} \cap {Y_{t'}} = \emptyset$ , when $t \ne {t'}$.The number of classes in each task may not be the same. The model learns one task at one time, and during each learning task, only the training data of the new task is available, while the data from past tasks $\left\{ {{x^1},...,{x^s}} \right\}$ is inaccessible in the new task. The model trained on the  classes of past tasks is defined as ${\varphi _{old}}(x;{\Theta _{old}})$. The goal of incremental learning is to train the classification model ${\varphi}(x;{\Theta})$ of the $t$-th task without suffering from catastrophic forgetting, where ${t>s}$, and ${x}$ represent the input samples and ${\Theta_{old}}$ and ${\Theta}$ respectively indicate model parameters.

\section{Scalable class-incremental learning based on parametric neural collapse}
\label{sec4}
The core objective of SCL-PNC is to design an incrementally expandable model backbone, in which the fully connected (FC) layer is replaced by a parameterized Equiangular Tight Frame (ETF) classifier for mitigating misalignment due to evolving class distributions, and through an adaptive feature transformation module construction, the extended model can effectively achieve neural collapse under the guidance of the ETF classifier for diminishing inter-module feature drift.

\subsection{Network architecture}
\label{subsec4.1}
\begin{figure}[t]
	\centering
	\includegraphics[width=1\textwidth,height=0.42\textheight]{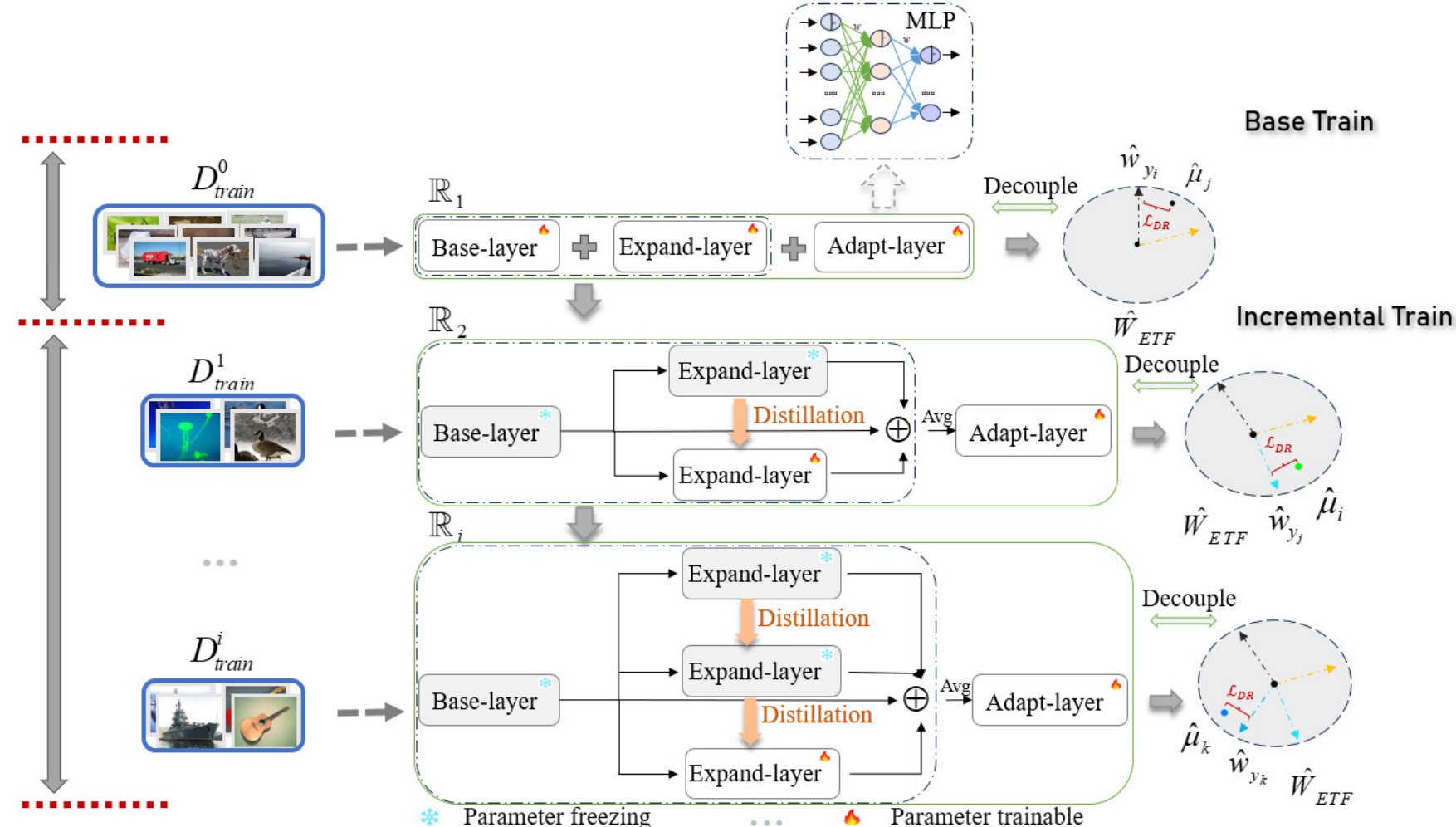}
	\caption{ Scalable class-incremental learning based on parametric neural collapse architecture}\label{Fig.2}
\end{figure}

The proposed model consists of three main components, which are an expandable model backbone, an adapt-layer, and a parametric ETF classifier. The expandable backbone includes a base-layer and multiple expand-layers. To retain knowledge from previous tasks, both the base-layer and expand-layers have their task-specific parameters frozen. The base-layer extracts stable representations and transmits them to the expand-layers, while inter-layer knowledge distillation preserves old-task continuity, alleviates overfitting to new classes, and mitigates knowledge discontinuity across model blocks. The adapt-layer aligns feature prototypes of the backbone with the classifier prototypes of corresponding categories, inducing neural collapse during incremental expansion. The parametric ETF classifier dynamically scales with the number of classes by extending its classifier vectors accordingly. The overall framework is illustrated in Fig.\ref{Fig.2}.

During training, the model strictly follows the class-incremental learning paradigm. Since SCL-PNC freezes parameters of previous-task modules when learning new tasks, the base training phase includes most classes to ensure robust and generalizable representations in the initial stage.

Formally, the proposed SCL-PNC framework consists of three major functional components: the base-layer $f_b( \cdot )$, a series of expand-layers $\{f_e^{(t)}(\cdot)\}_{t=1}^T$, and an adapt-layer $f_a( \cdot )$ followed by the parametric ETF classifier $g( \cdot )$.
For an input image $x \in \mathbb{R}^{H \times W \times C}$, the base-layer first extracts a general feature representation:
\[
\mu_b = f_b(x; \theta_b), \tag{1}
\]
where $\theta_b$ denotes the parameters of the frozen base-layer after initial training.

During the $t$-th incremental task, the newly added expand-layer $f_e^{(t)}$ refines both the base representation and the previous expand-layer's output:
\[
\mu_e^{(t)} = f_e^{(t)} \left( \left[ \mu_b, \mu_e^{(t-1)} \right] ; \theta_e^{(t)} \right), \tag{2}
\]
where $[\cdot,\cdot]$ denotes feature concatenation and $\theta_e^{(t)}$ are trainable parameters of the current expand-layer.

The adapt-layer then performs feature alignment and projection to the ETF classifier space:
\[
\mu_a^{(t)} = f_a\left( \mu_e^{(t)} ; \theta_a \right), \quad z^{(t)} = g\left( \mu_a^{(t)} ; W \right), \tag{3}
\]

where $W = [w_1, ..., w_K]$ represents the ETF classifier weights of $K$ classes.

To ensure inter-layer consistency, a knowledge distillation constraint is imposed between consecutive expand-layers:
\[\label{equ_4}
L_{distill}^{(t)} = \frac{1}{2} \left( \left( \hat{\mu}_e^{(t-1)} \right)^\top \hat{\mu}_e^{(t)} - 1 \right)^2, \tag{4}
\]
where $\hat{\mu}_e^{(t)}$ denotes the $L_2$-normalized feature vector.

The overall prediction process of SCL-PNC during the $t$-th incremental stage can be summarized as following.
\[
y^{(t)} = g\left( f_a\left( f_e^{(t)}  \left[ f_b(x), f_e^{(t-1)}(x) \right]  \right) \right), \tag{5}
\]

which explicitly shows the hierarchical and parallel flow of information from the base-layer to the dynamic expand-layers and the adapt-layer.

This mathematical formulation provides a clearer representation of the model’s functional dependencies and supports the experimental verification presented in Section 5.

In the experiment, the dataset's classes are divided into two subsets: the base train and the incremental train. The base train includes approximately half of the total categories, while the remaining categories are evenly distributed among the subsequent incremental tasks. During each incremental task, the classifier's class prototypes are dynamically expanded to accommodate the newly introduced classes.

In the base train phase, all model parameters are trainable. The model backbone is composed of a base-layer and a series of expand-layers. For the base task, the backbone includes one base-layer and one expand-layer. The model backbone can extract features from input images, while the adapt-layer guides the convergence of the network by minimizing a regression loss. This loss encourages the feature vectors to align with their corresponding classifier vectors located at the vertices of the ETF classifier, thereby promoting neural collapse.

During the first incremental phase, the parameters of all previously trained backbone modules are kept fixed. The newly added expand-layer receives inputs from both the preceding expand-layer and the general features extracted by the base-layer. Additionally, a knowledge distillation loss is applied between each pair of consecutive expand-layer to constrain the similarity of their outputs. This mechanism can mitigate knowledge inconsistency caused by distribution shifts between old and new tasks. Importantly, the adapt-layer remains trainable throughout the entire learning process to align the feature vector prototypes of the model backbone with the classifier prototypes of their respective categories, thereby fostering coherent and continuous feature representations across layers.

This architectural design effectively balances plasticity and stability, significantly alleviating feature drift during sequential expansions. It enhances the model’s ability to integrate new knowledge while preserving previously learned information, thus maintaining long-term memory.


\subsection{The Expandable Model Backbone}
\label{subsec4.2}
The backbone of SCL-PNC adopts an expandable architecture, which is demonstrated  in the dotted-wire frame of each task in Fig.\ref{Fig.2}. Specifically, the entire model backbone is divided into two main components, which are base-layer and expand-layers.  The fully trained and frozen base-layer is responsible for extracting general features from the input data, while a set of expand-layers is to capture task-specific information in an incremental learning manner.

During the base train phase, the model backbone comprises the base-layer and the first expand-layer. Through training on large-scale, multi-class data, the base-layer can learn the generalized and transferable representations, while the expand-layer further refines these features for enhancing discriminability. Under the guidance of a parametric ETF classifier and an adapt-layer, the model can achieve the strong classification performance by aligning feature vectors with their corresponding classifier prototypes.

\begin{figure}[t]
	\centering
	\includegraphics[width=0.55\textwidth,height=0.20\textheight]{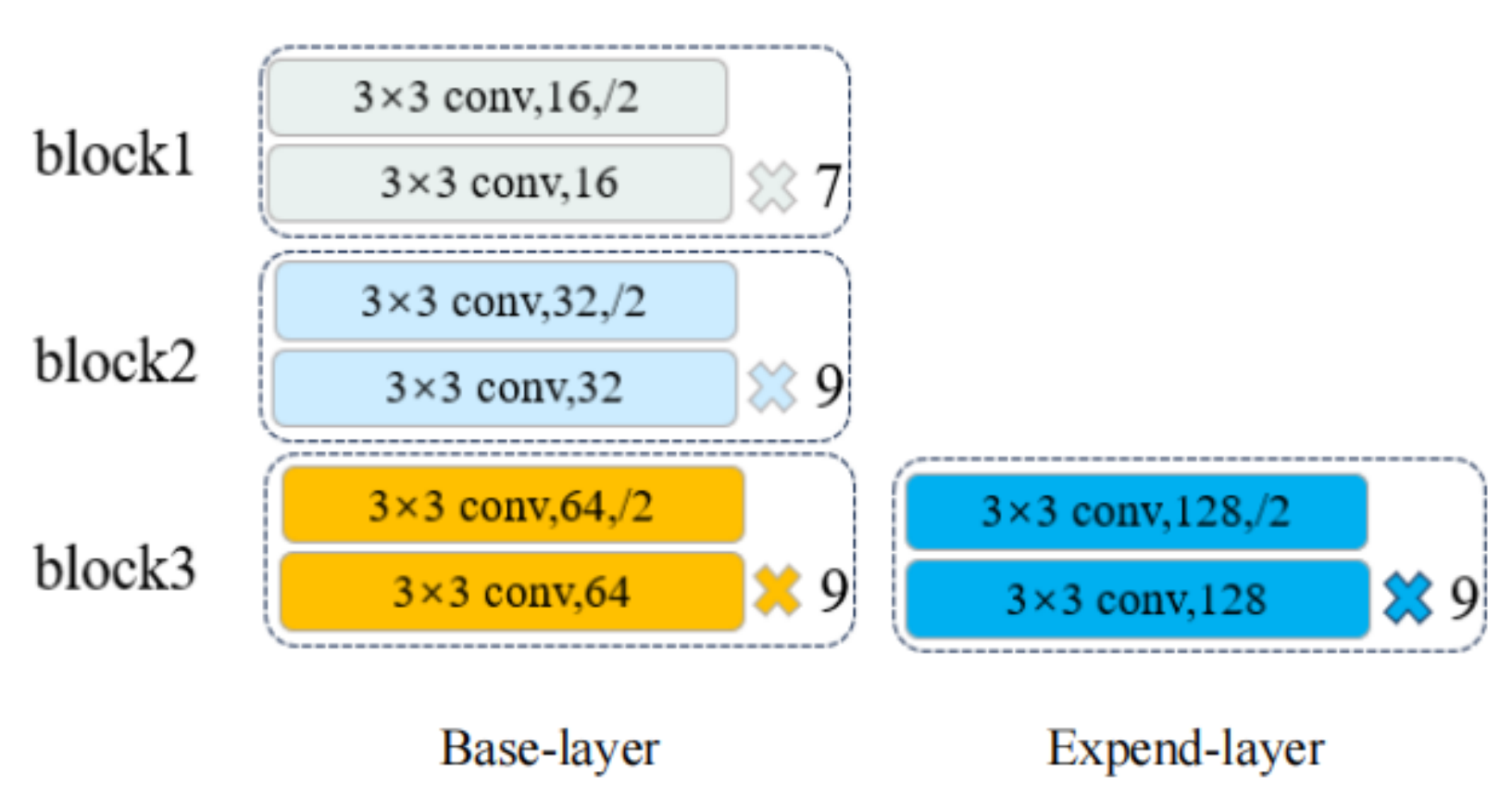}
	\caption{The structure of the base-layer and the expand-layer}\label{Fig.3}
\end{figure}

Compared to existing dynamic model expansion approaches, SCL-PNC can achieve the superior memory and the computational efficiency. Since the base-layer is frozen after initial training, there is no need to retrain its substantial parameters during incremental phases, thereby significantly reducing memory overhead. Although each incremental task introduces a new expand-layer, these layers are designed to be lightweight with compact parameterization. Through independent updates and distillation-based feature alignment, SCL-PNC can maintain the controlled parameter growth to effectively mitigate the inference efficiency degradation commonly caused by parameter explosion in long-term incremental learning scenarios. The design of the base-layer and the expand-layer draws inspiration from the residual block concept of the ResNet network, and is respectively the combination of the different block. The base-layer consists of the interconnection of three block from block1 to block3, while the expand-layer composes of one block in Fig.\ref{Fig.3}.

In order to further enhance representational stability during incremental backbone expansion, SCL-PNC adopts a parallel expansion architecture with knowledge distillation (P-KD) rather than the traditional serial expansion strategy. In a serial expansion configuration, each newly appended expand-layer relies solely on the output of its immediate predecessor, which leads to accumulated inter-module feature drift and information attenuation as the network depth increases. In contrast, the proposed parallel expansion structure enables every expand-layer to simultaneously receive general and stable features from the frozen base-layer and adaptive representations from the preceding expand-layer.

This design is grounded in the core assumption that the base-layer features act as anchor regularizers for deep-layer representations. The base-layer extracts transferable and stable low-level semantics that remain invariant across tasks, providing a consistent reference to constrain the feature evolution of subsequent modules. By using the base-layer output as an anchor and aligning deep representations through inter-layer knowledge distillation, the parallel expansion mechanism effectively suppresses feature drift and maintains long-term consistency across tasks.

Moreover, the P-KD framework achieves a better balance between stability (knowledge retention) and plasticity (new knowledge acquisition) by jointly optimizing the distillation loss (Equation \ref{equ_4}) and the point regression loss. The theoretical advantage of this architecture is further validated through feature similarity analysis in Section \ref{SA}.

\subsection{Adapt-layer}
\label{subsec4.4}
The direct integration of the ETF classifier with the designed expand-layer can not yield satisfactory results because of the inter-module feature drift of the different layers. Inspired by C-FSCIL\cite{hersche2022constrained}, we construct an adapt-layer to guide the features from the model backbone to collapse toward the vertices of the ETF classifier. 

Specifically, the adapt-layer can constrain the feature vector prototypes from the model backbone to align with the classifier prototypes corresponding to each category during incremental expansion, thereby promoting neural collapse within the feature of the model backbone. Furthermore, the adapt-layer can serve as a bridge of feature space transform between the model backbone and the parametric ETF classifier to eliminate the inter-module feature drift of the different layers. Accordingly, the number of neurons in the first layer of the adapt-layer matches the dimension of the backbone’s output features, while the number of neurons in the final layer aligns with the input dimension of the ETF classifier.

The multilayer perceptron (MLP) is a simpler method for high computational efficiency and convergence stability. Therefore, we select the MLP-based design for the adapt-layer. In experiments, we specifically analyze other option for adapt-layer. 
Formally, given an input sample $x$ with backbone feature $\mu = f(x; \theta_f) \in \mathbb{R}^d$, the adapt-layer transforms it into the classifier space through a multilayer perceptron (MLP):

\[
z = g(\mu; \theta_a), \tag{9}
\]

where $g(\cdot; \theta_a)$ denotes the MLP-based mapping function parameterized by $\theta_a$, and $z \in \mathbb{R}^{d'}$ is the adapted feature representation. The first layer of the MLP matches the dimension of the backbone output, while the last layer aligns with the input dimension of the ETF classifier.

To guide the adapted feature $z$ to collapse toward the vertices of the ETF classifier, we follow the definition of the point regression loss and reformulate it to constrain the adapt-layer output.

\[
L_{DR}(z, w_k) = \frac{1}{2\sqrt{E_W E_Z}} \left( w_k^\top z - \sqrt{E_W E_Z} \right)^2, \tag{10}
\]

where classifier weight vectors $w_c$ represents the ETF prototype corresponding to class $k$ ($w_k \in W$, $1\le k\le K$, $W$ is classifier weight matrix), and is predefined by the function of the class number $K$ in Equation (\ref{define_w}).  $E_W$ and $E_Z$ denote the $l_2$ norm constraints of the classifier vector $w_c$ and the feature vectors $z$ respectively.
This loss encourages $z$ to align with its corresponding class prototype, thereby reinforcing neural collapse in the adapted feature space.







\subsection{Parametric ETF Classifier}
\label{subsec4.5}
It is well established that the core objective of classification tasks is to minimize the feature distance among samples belonging to the same class while maximizing the feature distance between samples from different classes. The phenomenon of neural collapse can follow the above rule. In the final stages of model training—particularly when the training loss approaches zero, the highly compact intra-class features and the well-separated inter-class features of the final layer in the neural network can align with their corresponding classifier weight vectors. Ultimately, the features collapse to a set of equiangular directions, forming the vertices of an ETF. At this stage, the feature representations of all classes follow some uniform distributions in equiangular space,which collectively constitute the structure of an ETF classifier.

The geometric optimality of the feature space based on equiangular directions is manifested through a synergistic optimization mechanism, which minimizes the within-class covariance matrix  while maximizing the between-class covariance matrix . This optimization objective is equivalent to the global maximization of the Fisher discriminant criterion, which achieves the theoretically optimal solution for linear separability via the extremal analysis.

The vertex distribution properties of an ETF can provide a complete geometric characterization of such optimal solutions in high-dimensional space. The equiangular symmetry of the ETF ensures the dual optimization of the inter-class separability and the intra-class compactness. It follows that the optimal classifier should inherently adopt an ETF structure.

Specifically, the classifier is initialized as an ETF, the classifier matrix $W$ of which maintains a fixed magnitude $\sqrt{E_W}$. During training, only the feature representations $Z$ are optimized. Following the layer-peeled model, the ETF classifier can optimize the feature representation by minimizing the loss $L_{DR_T}$ as following.

\begin{equation}\label{eq:optimization_problem}
	\begin{cases}
		\displaystyle
		\min_{Z}L_{DR_T}=\min_{Z} \frac{1}{Kn_{k}} \sum_{k=1}^K \sum_{i=1}^{n_k} L_{DR}(z_{k,i}, w_k), \\
		\text{s.t.} \quad \|z_{k,i}\|^2 \le E_Z, \quad \forall 1 \le k \le K, \ 1 \le i \le n_k.
	\end{cases}
	\tag{13}
\end{equation}
Where, ${W}$ denotes a unitary equiangular ETF classifier, ${K}$ represents the total number of classes in the training task, $z_{k,i}$ is the feature vector of the $i$th sample of the $k$th class ($z_{k,i}\in Z$), $n_k$ indicates the sample number of the $k$th class, and $L_{DR_T}$ signifies the point regression loss of all samples. The matrix ${W}=[w_1,w_2,...,w_K]$ must satisfy the following conditions:

\begin{equation}\label{Eq.3}
	w_i^T w_j = E_W \left(\frac{K}{K-1} \delta_{i,j} - \frac{1}{K-1}\right), \quad \forall i,j \in [1,K],
	\tag{14}
\end{equation}

Where, ${\delta_{i,j}}$ serves as an indicator function, which takes the value of 1 when ${i=j}$ and 0 otherwise.
Before enforcing the equiangular constraint in Eq.(14), we explicitly construct the ETF classifier prototypes.

The classifier weight matrix $W_t = [w_1, ..., w_{K_t}]$ of $t$th incremental task is generated by projecting projecting the vertices of a $K_t^{-1}$-simplex onto  onto a centered subspace:
\[\label{define_w}
w_k = \sqrt{E_W} \left( e_k - \frac{1}{K_t} \mathbf{1} \right), 1\le k\le K_t,\tag{15}
\]
where $e_k$ is the canonical basis vector of the $k$th class and $\mathbf{1}$ is an all-one vector. This construction guarantees that all class vectors are equiangular and have equal norm. $w_k$ is parameterized by $K_t$, and  defines the class weight vector of parametric ETF classifier. When new classes appear, additional simplex vertices are appended with the class number $K_t$ changing, allowing the classifier to expand dynamically while preserving the ETF geometry.

\subsection{Model Loss}
To preserve the feature consistency between consecutive incremental stages, we construct the total distillation Loss according to Eq. (\ref{equ_4}):

\[L_{distill_T} = \sum_{t=1}^{T}L_{distill}^{(t)}  \tag{11}\]

Finally, the model is optimized with the following the total loss.

\[\label{total_loss} L_{total} = L_{DR_T} + \lambda L_{distill_T}   \tag{12}\]

where $\lambda$ controls the trade-off between feature alignment and distillation consistency.

\section{Experiments}
\label{sec5}
\subsection{Experimental Setup}
\label{subsec5.1}
To evaluate the performance of the proposed SCL-PNC, we conduct experiments on the CIFAR-100\cite{krizhevsky2009tiny} and ImageNet-100\cite{deng2009imagenet} datasets. Specifically, the CIFAR-100 dataset is a labeled subset from the 80 Million Tiny Images dataset, and each category includes 600 images (500 images for training and 100 images for testing) in  table \ref{tab:dataset_summary}. In contrast, ImageNet-100 is a subset of the ImageNet-1000 benchmark, and each category contains about 1300 images for training and 50 images for testing in  table \ref{tab:dataset_summary}. In B50Inc10, B represents the base task, the number following B indicates the category count, Inc stands for the incremental task, the number following Inc denotes the category count, and others have similar meanings. Experiments employ PyTorch and the PyCIL library to benchmark class-incremental learning methods under identical hardware conditions. For some methods requiring data replay (e.g., rehearsal-based approaches), a fixed buffer size of 2,000 exemplars is allocated, while non-replay methods (e.g., LwF) are evaluated without exemplar storage. In model training, the initial learning rate set to 0.1, the momentum of SGD is set to 0.9, the training epoch is 200, the batch size is 128 in the base task and the incremental task, and the learning rate of each round decays by 0.01 after the 20 training round. The data augmentation method techniques include random cropping, horizontal flipping, and color enhancement.

\begin{table}[h]
    \centering
    \caption{Summary of dataset statistics and partitioning strategies for CIFAR-100 and ImageNet-100}
    \label{tab:dataset_summary}
    \resizebox{\textwidth}{!}{%
    \begin{tabular}{l c c c c c c}
        \toprule
        \textbf{Dataset} & 
        \textbf{\makecell{Total \\ Training \\ Images}} & 
        \textbf{\makecell{Total \\ Testing \\ Images}} & 
        \textbf{\makecell{Total \\ Classes}} & 
        \textbf{Scenario} & 
        \textbf{\makecell{Base \\ Classes \\ /Images}} & 
        \textbf{\makecell{Incremental \\ Classes \\ /Images per \\ Task}} \\
        \midrule
        CIFAR-100 & 50,000 & 10,000 & 100 & B50Inc10 & 50/30,000 & 10/6,000 \\
        & & & & B50Inc5 & 50/30,000 & 5/3,000 \\
        & & & & B40Inc3 & 40/24,000 & 3/1,800 \\
        & & & & B10Inc10 & 10/6,000 & 10/6,000 \\
        \midrule
        ImageNet-100 & 130,000 & 5,000 & 100 & B50Inc10 & 50/67,500 & 10/13,500 \\
        & & & & B50Inc5 & 50/67,500 & 5/6,750 \\
        & & & & B50Inc2 & 50/67,500 & 2/2,700 \\
        \bottomrule
    \end{tabular}%
    }
\end{table}

\subsection{Evaluation Metric}
\label{subsec5.2}
The evaluations of the incremental learning systems adopt the incremental accuracy curve tracks the model evolution capability to assimilate new knowledge across sequential tasks, thereby reflecting its adaptability over time. The method can capture the critical trade-off between the new knowledge acquisition and the previous information retention —an inherent challenge in the incremental learning. The key quantitative metric of the incremental accuracy curve is average recognition accuracy, which measures overall performance across all tasks. The metric offers a rigorous and multi-dimensional evaluation of a model capacity for the incremental learning while providing valuable reference for algorithm design and optimization in open-ended learning scenarios.


Average Incremental Accuracy ($\text{Acc}_{avg}$) is the primary metric for measuring the overall performance of the model across all completed tasks. It computes the average recognition accuracy achieved on the union of all seen classes after the completion of each incremental task.

\[
\text{Acc}_{avg} = \frac{1}{T} \sum_{t=1}^T A_t \tag{16}
\]
Where, $T$ is the total number of incremental tasks.
$A_t$ is the classification accuracy on the test set of all classes seen from task 1 up to task $t$, after the model has been trained on task $t$. A higher $\text{Acc}_{avg}$ indicates superior overall performance.

\subsection{Comparative Analysis on Small-Scale CIFAR-100 Datasets}
This section presents a systematic evaluation of the various class incremental learning methods on the small-scale CIFAR-100 dataset, aiming to investigate their learning efficacy and generalization capabilities under data constrained conditions. A comparative analysis was conducted across different incremental learning strategies to assess method performance from the different situations. The main difference of these strategies are the class number of the base task and the incremental task, and include B50Inc10, B50Inc5, B10Inc10, and B40Inc3. 

To validate our central hypothesis(when employing the parameter-freezing method of the model backbone, it is essential to include a large number of classes in the base task to facilitate the acquisition of the stable and generalizable feature representations.), we conduct the classification experiments of SCL-PNC of the different incremental learning strategies on CIFAR-100 dataset. Table \ref{Tab.1} details the per-task accuracy and the average accuracy of all tasks of SCL-PNC under distinct experimental strategies. Experimental results show that  SCL-PNC initialized with a base task containing a substantial number of classes exhibits lower forgetting rates and maintains the more stable performance across subsequent incremental phases. In contrast, when the base task includes only 10 classes, SCL-PNC achieves the high initial accuracy but experiences the significant performance degradation in later stages, accompanied by the sharp increase in forgetting rates and the compromised feature representation stability. In table \ref{Tab.1}, the number 0 denotes the base task, while numbers 1 to 10 represent the incremental tasks.

\begin{table}[ht]
\caption{Classification performance of SCL-PNC of the different incremental learning strategies on CIFAR-100 dataset, $t$ is the serial number of each incremental task }\label{Tab.1}
\resizebox{\textwidth}{!}{%
\begin{tabular}{c|*{11}{c@{\hspace{6pt}}}|c}
	\hline
	\textbf{Strategy} & $t$=0 & $t$=1 & $t$=2 & $t$=3 & $t$=4 & $t$=5 & $t$=6 & $t$=7 & $t$=8 & $t$=9 & $t$=10 &  $\text{Acc}_{avg}$ \\
	\hline
	B50Inc10 & 78.62 & 74.67 & 73.04 & 68.20 & 66.28 & 64.69 & -- & -- & -- & -- & -- & 70.92 \\
	B50Inc5 & 78.62 & 75.76 & 74.80 & 73.02 & 72.61 & 70.31 & 68.03 & 67.55 & 66.80 & 66.33 & 65.21 & 70.82 \\
	B10Inc10 & 91.10 & 79.40 & 74.37 & 68.45 & 64.38 & 59.93 & 57.51 & 52.74 & 50.66 & 50.48 & -- & 64.90 \\
	\hline
\end{tabular}%
}
\end{table}

To compare the performance of the proposed SCL-PNC with the state-of-the-art methods, experiments benchmark the average classification accuracy of the per-task, along with the corresponding model parameter counts under the B50Inc10 strategy. These state-of-the-art methods contains LwF\cite{8107520}, iCaRL\cite{rebuffi2017icarl}, WA\cite{zhao2020fair}, DER \cite{yan2021der}, FOSTER\cite{foster2022feature}, MEMO\cite{zhou2022memory}, BEEF \cite{wang2022beef}, DS-AL\cite{zhuang2024dsal}, and FCS\cite{li2024fcs}.  Specifically, LwF\cite{8107520} can uses only new task data to train the network while preserving the original capabilities. iCaRL\cite{rebuffi2017icarl} can learning the dynamic data representation to be compatible with deep learning architectures. WA\cite{zhao2020fair} can corrects the biased weights in the FC layer after normal training process by weight aligning. DER\cite{yan2021der} can dynamically expand the representation according to the complexity of novel concepts by introducing a channel-level mask-based pruning strategy. FOSTER\cite{foster2022feature} can gradually fit the residuals between the target model and the previous ensemble mode for empowering the model to learn new categories adaptively. MEMO\cite{zhou2022memory} can extract diverse representations with modest
cost and maintain representative exemplars by memory-efficient expandable model. BEEF \cite{wang2022beef} can decouple modules to achieve bi-directional compatibility, and integrate them into a unifying classifier with minimal cost to alleviate the conflicts among modules. DS-AL\cite{zhuang2024dsal} can contain a main stream offering an analytical linear solution, and a compensation stream improving the inherent under-fitting limitation due to adopting linear mapping by exemplar-free CIL setting. FCS\cite{li2024fcs} can adapt prototypes of old classes to the new model by feature calibration network and enhance feature separation among different classes by prototype-involved contrastive loss. However, these method ignore the feature drift between modules in term of expand module and the misalignment due to evolving class distributions because of the classification boundaries reallocation of the incremental data. Therefore, the proposed SCL-PNC enables demand-driven, minimal-cost backbone expansion by adapt-layer and refines the static into a dynamic parametric Equiangular Tight Frame (ETF) framework according to incremental class to solve the above question. As shown in table \ref{Tab.2}, SCL-PNC achieves the highest average recognition accuracy compared to all other approaches with the acceptable parameter count. In the table, bold numbers indicate the highest recognition accuracy, underlined values represent the second-best performance, and ${\#P}$ denotes the number of model parameters (in millions). 

Although SCL-PNC performs slightly worse than the FCS\cite{li2024fcs} approach during the base task and the first incremental task, it demonstrates superior overall performance in subsequent incremental tasks. Specifically, compared to the classification performance after the base task, SCL-PNC experiences only a ${3.59\%}$ drop in accuracy during the first incremental task, whereas the FCS method suffers a ${9.07\%}$ decline. Starting from the second incremental task, SCL-PNC consistently achieves the best classification accuracy among all methods, primarily due to its significantly lower performance degradation. Overall, SCL-PNC demonstrates the highest average recognition accuracy and the most stable performance across the entire incremental learning process. The main reason is that the prototype-involved contrastive loss of FCS is effective in the initial incremental learning, but FCS can not adapt the class redistribution with the old class forgetting.

\begin{table}[ht]
	\centering
	\caption{Experimental results of B50Inc10 on the CIFAR-100 dataset, $t$ is the serial number of each incremental task}
	\resizebox{\textwidth}{!}{%
		\begin{tabular}{lccccccccc}
			\toprule
			\textbf{Method} & \textbf{\#P(M)} & $t$=0 &$t$=1 & $t$=2 & $t$=3 & $t$=4 &
		  $t$=5 &  $\text{Acc}_{avg}$ & PD \\
			\midrule
			LwF (PAMI’18) \cite{8107520}     & 0.47  & 76.0 & 45.68 & 35.2 & 26.88 & 24.5 & 23.17 & 38.57 & 52.83 \\
			iCaRL (CVPR’17) \cite{rebuffi2017icarl}     & 0.46  & 75.34 & 65.48 & 62.01 & 55.85 & 53.02 & 50.49 & 60.37  & 24.85 \\
			WA (CVPR’20) \cite{zhao2020fair}        & 0.46  & 75.92 & 68.63 & 65.70 & 61.50 & 58.80 & 56.81 & 64.56  & 19.11\\
			DER (CVPR’21) \cite{yan2021der}       & 9.27  & 76.50 & 72.78 & 71.40 & 68.14 & 65.84 & 64.10 & 69.79  & \textbf{12.40}\\
			FOSTER (ECCV’22) \cite{foster2022feature}    & 0.46  & 78.02 & 72.30 & 69.96 & 63.86 & 63.04 & 59.66 & 67.81 & 18.36 \\ 
			MEMO (ICLR’23) \cite{zhou2022memory}     & 7.14  & 76.30 & 67.48 & 66.19 & 62.35 & 60.01 & 58.12 & 65.08  & 18.18\\
			BEEF (ICLR’23) \cite{wang2022beef}      & 2.30  & 77.96 & 70.82 & 69.83 & 63.30 & 60.63 & 59.48 & 67.00 & 18.48\\
			DS-AL (AAAI’24) \cite{zhuang2024dsal}     & 0.46  & 77.94 & 71.15 & 67.51 & 63.55 & 60.67 & 58.43 & 66.54  & 19.53 \\
			FCS (CVPR’24) \cite{li2024fcs}       & 12.20 & 83.92 & 74.85 & 70.77 & 67.04 & 63.66 & 62.13 & 70.40  & 21.79 \\
			PRL(NIPS’24) \cite{shi2024prospective}     & -  & 82.80 & 75.65 & 72.10 & 68.26 & 65.52 & 63.44 & \underline{71.29 }  & 19.36 \\
			TagFex(CVPR’25) \cite{zheng2025task}     & -  & - & - & - & - & - & 70.33 & \textbf{75.87} & - \\
			SCL-PNC                        & 8.54  & 78.62 & 74.67 & 73.04 & 68.20 & 66.28 & 64.69 & 70.92  & \underline{13.93} \\
			\bottomrule
		\end{tabular}%
	}
	\label{Tab.2}
\end{table}

\begin{figure}[htbp]
	\centering
	\begin{subfigure}[b]{\textwidth}
		\includegraphics[width=\textwidth]{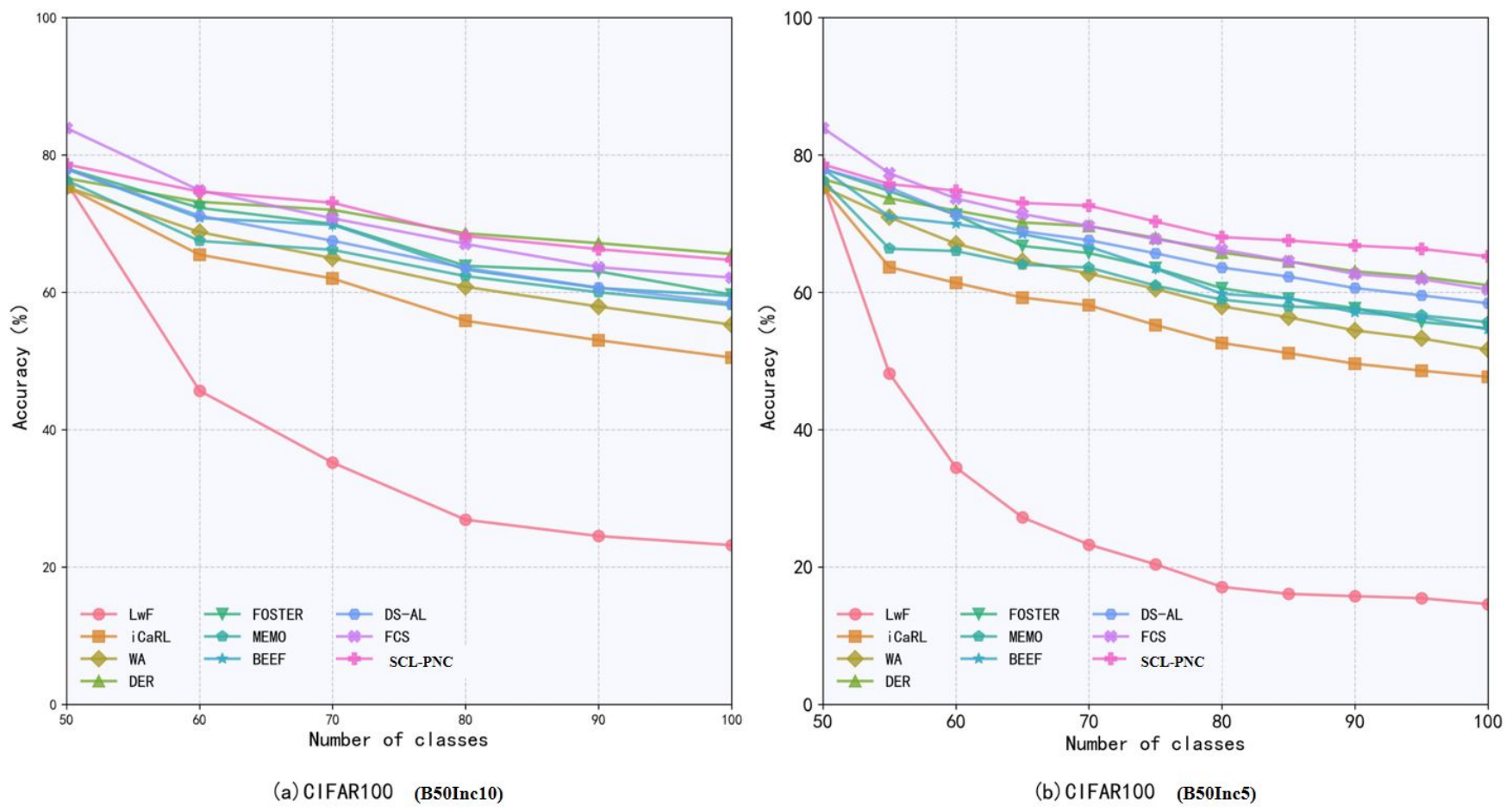}
		\vspace{-2mm}  
	\end{subfigure}
	\begin{subfigure}[b]{\textwidth}
		\includegraphics[width=\textwidth]{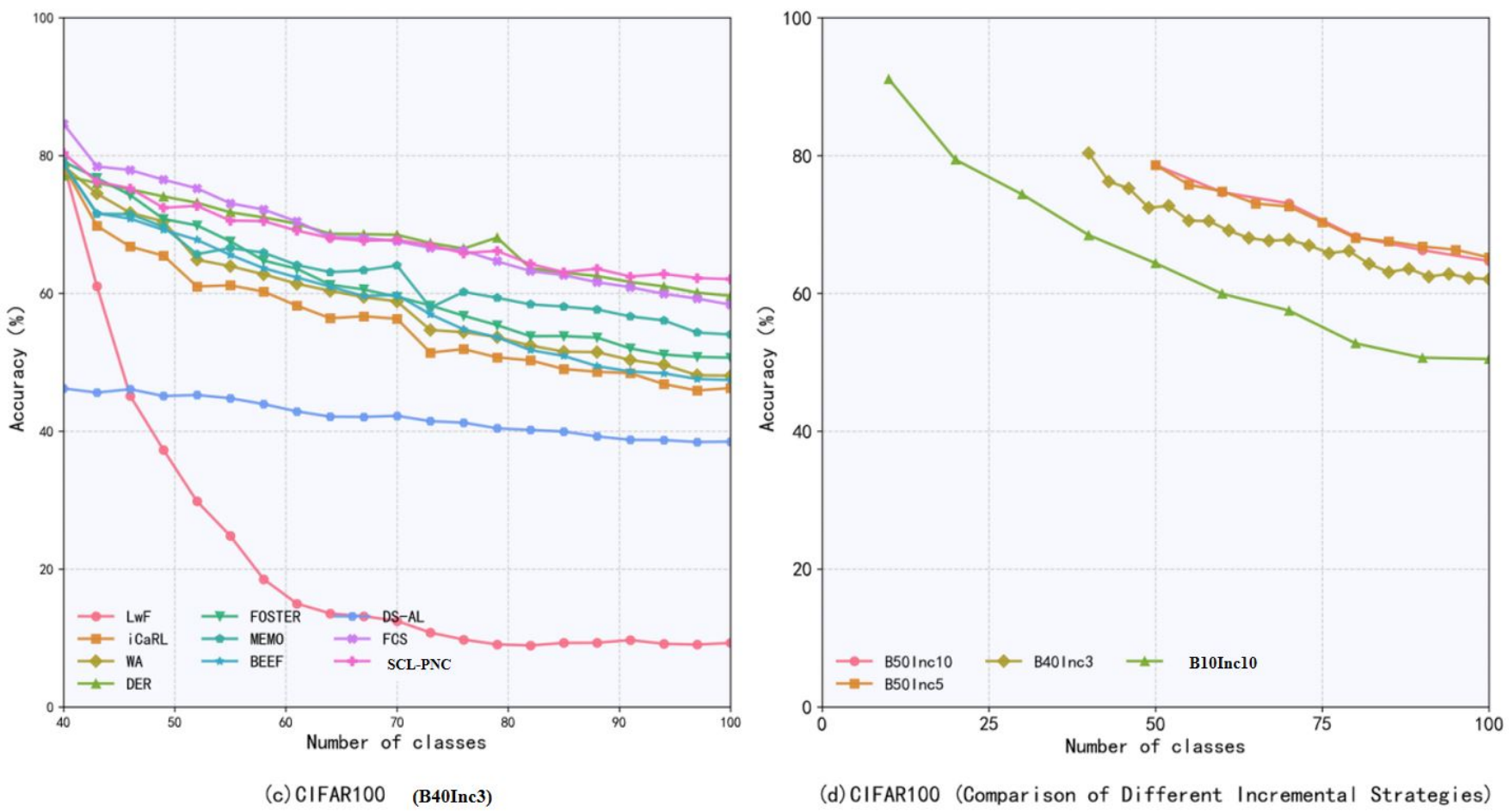}
	\end{subfigure}
	\caption{Average Accuracy of the Class Incremental Learning Methods Under Different Experimental Strategies on the CIFAR-100 Dataset}
	\label{Fig.4}
\end{figure}
To evaluate the classification accuracy trajectories of the different methods, we conduct the detail experiments of the various strategies, which include B50Inc10,B50Inc5, B40Inc3 and B10Inc10. The performance curves in Fig.\ref{Fig.4} demonstrate the superior classification accuracy of SCL-PNC across various incremental learning strategies on the CIFAR-100 dataset. In Fig.\ref{Fig.4}(a), the B50Inc10 strategy allocating the majority of classes to the base task enables the model to acquire more stable and generalizable feature representations.
Fig.\ref{Fig.4}(b) presents the results of the more challenging B50Inc5 strategy, which imposes the stricter demands on the model ability to resist forgetting due to the finer-grained incremental updates. The extreme low-increment scenario shown in Fig.\ref{Fig.4}(c), and the experimental results further demonstrates the robustness of SCL-PNC in long-term incremental learning. Notably, Fig.\ref{Fig.4}(d) illustrates that SCL-PNC maintains the stronger performance stability as the number of the base class increases, exhibiting significantly reduced accuracy degradation. This consistent performance improvement across different experimental strategies confirms the SCL-PNC to learn stable representations and to integrate new knowledge while preserving previously acquired information. These results particularly underscore the method effectiveness in maintaining representational stability and mitigating catastrophic forgetting, especially under settings involving the larger class increments of the more base class.

Table \ref{Tab.3} systematically compares the average accuracy of class-incremental learning methods under different experimental strategies, which contain the different incremental learning tasks B50Inc10 (T=5), B50Inc5 (T=10) and B40Inc3 (T=20), in which T indicates the number of the incremental task. The results in the table demonstrate that, across different experimental settings, SCL-PNC achieves substantially better performance than both prior dynamic network methods and traditional exemplar replay-based approaches. Notably, the performance advantages are maintained consistently regardless of the task sequence length (T=5/10/20) or the base class quantity, highlighting SCL-PNC robustness in handling varying degrees of incremental learning complexity. This superior performance can be attributed to the SCL-PNC of the effective parallel knowledge integration mechanism, the preservation ability of feature stability across tasks through distillation constraints and parametric ETF classifier.

\begin{table}[!ht]
	\centering
	\caption{Average accuracy under different incremental steps on the CIFAR-100 dataset (T denotes the number of the incremental task)}
	\label{Tab.3}
	\small
	\setlength{\tabcolsep}{6pt}
	\begin{tabular}{>{\raggedright\arraybackslash}p{3.6cm} c c c}
		\toprule
		\textbf{Method} & \textbf{B50Inc10 (T=5)} & \textbf{B50Inc5 (T=10)} & \textbf{B40Inc3 (T=20)} \\
		\midrule
		LwF (PAMI'18) [40]    & 38.57 & 28.04 & 21.13 \\
		iCaRL (CVPR'17) [1]   & 60.37 & 56.60 & 56.19 \\
		WA (CVPR'20) [30]      & 64.56 & 59.46 & 59.10 \\
		DER (CVPR'21) [5]     & 69.79 & 67.86 & 67.83 \\
		FOSTER (ECCV'22) [13]  & 67.81 & 64.35 & 61.15 \\
		MEMO (ICLR'23) [14]    & 65.08 & 62.18 & 62.71 \\
		BEEF (ICLR'23) [31]    & 67.00 & 64.03 & 59.03 \\
		DS-AL (AAAI'24) [32]   & 66.54 & 66.46 & 42.05 \\
		FCS (CVPR'24) [33]     & 70.40 & 69.04 & 68.36 \\
		PRL(NIPS’24) \cite{shi2024prospective}  & 71.26 & \underline{70.17 }& \textbf{68.44}\\
		TagFex(CVPR’25) \cite{zheng2025task} & \textbf{75.87} & - & - \\
		\textbf{SCL-PNC}    & \underline{70.92} & \textbf{70.82} & \underline{68.43} \\
		\midrule
		\multicolumn{4}{c}{\textbf{B100Inc0 (T=0)}} \\
		\midrule
		SCL-PNC (Joint Training) & \multicolumn{2}{c}{\textbf{73.07}} & \\
		\bottomrule
	\end{tabular}
\end{table}
As shown in table \ref{Tab.3}, under the B100Inc0 scenario, the SCL-PNC (Joint Training) variant reaches a performance of 73.07. Note that this baseline is not an incremental learning method but is included only as an upper-bound reference to indicate the theoretical performance limit when all data are accessible at once.

The comparative results clearly demonstrate that SCL-PNC achieves the superior experimental performance when benchmarks against the dynamic network-based approaches (e.g., DER, FOSTER, MEMO). These results substantiate that the selective expansion of the deep layers can effectively preserve model plasticity while simultaneously mitigating catastrophic forgetting. Furthermore, the experimental results across different strategies reveal a critical insight, which is that  the majority of classes incorporation during the base task phase significantly enhances the model's capability to learn more robust and generalizable representations. These observations collectively highlight two key advantages of SCL-PNC. One is its parameter-efficient architecture design of the parallel expand-layer can balance the plasticity and the stability of model, and the other is its adaptability of the adapt-layer can align the initial and incremental class distribution for optimal representation learning.

To further substantiate the capability of the proposed SCL-PNC method in achieving robust representational stability and adaptability against catastrophic forgetting, the evolution of the feature embedding space across consecutive incremental tasks ($t_i \rightarrow t_{i+1}$) was visualized using t-SNE, as presented in Figure 5.

In the provided visualizations, the feature distributions of old classes before the current task are denoted by gray points ($\circ$). Colored triangles ($\bigtriangleup$) represent the newly introduced classes after training for task $t_{i+1}$. Crucially, the colored arrows ($\rightarrow$) illustrate the drift trajectory of old-class features, and it's scale quantifying the vector from the old class centroid at $t_i$  to the updated centroid at $t_{i+1}$ .

\begin{figure*}
    \centering
     \includegraphics[width=1.05\textwidth]{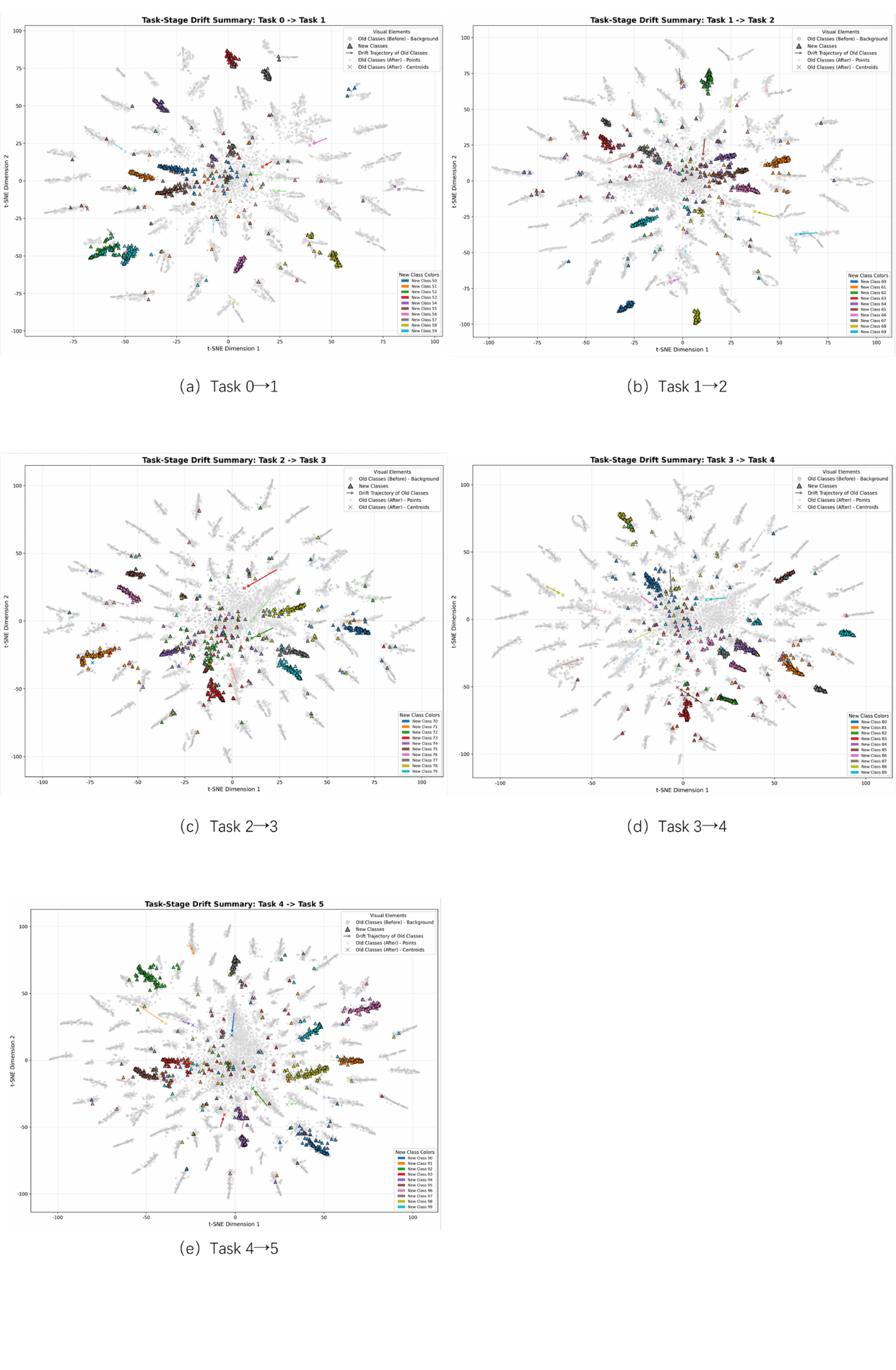}

    \vspace{-50pt}
    
    \caption{Task-stage drift on CIFAR-100 feature space via t-SNE. The visualization tracks feature space evolution from task $0 \to 1$ (a) to task $4 \to 5$ (e). Gray points show old-class features before the task; colored triangles show new-class features after the task. Colored arrows denote the minimal drift trajectory of old-class centroids, confirming representational stability.}
    \label{fig:t-sne-drift}
\end{figure*}
The visual evidence clearly demonstrates the SCL-PNC's ability to maintain a critical balance between stability and plasticity. In the early incremental stages (Figure 5a and 5b), the features of the newly added classes ($\bigtriangleup$) are observed to form compact and highly distinguishable clusters, occupying distinct, non-overlapping regions relative to the existing old-class manifold. This confirms the model's high plasticity and efficient assimilation of new knowledge. Simultaneously, the markedly short length of the old-class drift trajectories ($\rightarrow$) provides direct evidence of the model’s intrinsic stability.

As the incremental process continues (Figure 5c-e), a consistent pattern of minimal feature drift is maintained. The lengths of the old-class drift trajectories remain relatively minor throughout the sequence, indicating that the previously learned discriminative features are only marginally adjusted to incorporate new class boundaries, rather than being fundamentally restructured or catastrophically overwritten. Even in the final stage (Figure 5e), the old-class features retain a clear spatial organization, and the final set of new classes successfully establishes well-clustered, non-mixed locations within the overall embedding space.

Collectively, these qualitative analyses provide strong empirical validation that SCL-PNC achieves a robust stability-plasticity trade-off. The consistent preservation of old-class feature topology, coupled with the efficient embedding of new categories, directly supports the model's superior performance in continual learning benchmarks.

\subsection{Comparative Analysis on Large-Scale Dataset}
\label{subsec5.4}

To evaluate the performance of SCL-PNC on large-scale datasets, we conduct a comprehensive set of experiments, focusing on average incremental accuracy on the ImageNet-100 dataset. We strictly adhere to the standard CIL protocol for all comparisons (excluding LwF) and directly cite published results where experimental settings are identical.
As shown in Table \ref{Tab.4}, a common challenge observed across all methods is the decreasing trend in average accuracy as the number of incremental tasks increases, a direct consequence of catastrophic forgetting.

In the B50Inc10 ($T=5$) setting, $\text{BEEF}$ achieve the highest accuracy, leveraging a prior distribution to constrain the parameter space. However, BEEF incur the longest training time, several times longer than SCL-PNC. 

In the challenging B50Inc2 ($T=25$) setting, $\text{DS-AL}$ obtain the highest score, benefiting from a computationally expensive active sampling strategy (calculating uncertainty/diversity scores for all unlabeled samples). SCL-PNC’s strong performance in this long-horizon setting is attributed to the dynamic parametric ETF classifier, which effectively mitigates the misalignment due to evolving class distributions problem caused by the continuous, small-step increase in class categories, ensuring robust feature alignment across 25 tasks. Furthermore, we rigorously assessed the computational efficiency against $\text{DS-AL}$ (analytical sampling). DS-AL's requirement to compute uncertainty or diversity metrics via a complete forward pass over the entire dataset at every incremental step incurs approximately $10\times$ the computational cost of standard incremental training on ImageNet-100. By sharp contrast, SCL-PNC avoids sample-wise uncertainty estimation and restricts optimization to newly introduced classes, thereby reducing the total number of forward passes by $\approx 90\%$, achieving significant efficiency gains without compromising performance.

In the B50Inc5 ($T=10$) scenario, SCL-PNC achieve a competitive performance of $75.52\%$. The leading method, $\text{DER}$, achieves its peak accuracy at the expense of significantly increased model complexity, as it expands an entire subnetwork for task isolation. Specifically, $\text{DER}$(dynamic structural expansion) allocates a full subnetwork per task, leading to a substantial parameter count of 9.27M, severely limiting its scalability. Conversely, SCL-PNC employs a lightweight and principled expansion mechanism, adding only minimal adapt-layers and parametric ETF classifiers. This results in a compact architecture with 8.54M parameters, representing an approximate 8\% reduction relative to DER. This parameter efficiency is key to SCL-PNC's claim of scalability. Although the performance of FOSTER and MEMO is better in the B50Inc5 ($T=10$) scenario, they suffer from the decline performance in the B50Inc10 ($T=5$) or B50Inc2 ($T=25$) scenario. These results shows the adaptability of $\text{SCL-PNC}$ in the different scenarios.  $\text{SCL-PNC}$’s competitive performance, without this severe increase in model size, highlights the efficacy of its progressive layer freezing and parametric ETF classifier in balancing knowledge stability and representational capacity. 

\begin{table}[t]
	\centering
	\caption{Average accuracy under different incremental settings on the ImageNet-100 dataset (T denotes the number of the incremental task)}
	\resizebox{\textwidth}{!}{%
		\begin{tabular}{lccc}
			\toprule
			\textbf{Method} & \textbf{B50Inc10 (T=5)} & \textbf{B50Inc5 (T=10)} & \textbf{B50Inc2 (T=25)} \\
			\midrule
			LwF (PAMI'18) \cite{8107520}   & 46.24 & 7.60  & -- \\
			iCaRL (CVPR'17) \cite{rebuffi2017icarl}   & 62.62 & 59.56  & 54.56 \\
			WA (CVPR'20) \cite{zhao2020fair}      & 65.81 & 63.71  & 58.34 \\
			DER (CVPR'21) \cite{yan2021der}     & \underline{77.42  }   & \textbf{77.73}  & --     \\
			FOSTER (ECCV'22) \cite{foster2022feature}  & 70.01    & \underline{77.54}  & 69.34 \\
			MEMO (ICLR'23) \cite{zhou2022memory}    & 76.83 & 77.27  & --     \\
			BEEF (ICLR'23) \cite{wang2022beef}    & 77.27 & --      & --     \\
			DS-AL (AAAI'24) \cite{zhuang2024dsal}   & 75.20 & --      & \textbf{75.03} \\
			FCS (CVPR'24) \cite{li2024fcs}     & 74.06 & 52.43  & --     \\
			PRL(NIPS’24) \cite{shi2024prospective} & 72.85 & 71.54  & --     \\
			TagFex(CVPR’25) \cite{zheng2025task} & \textbf{80.64} & -  & --     \\
			\textbf{SCL-PNC}    & 76.80 & 75.52 & \underline{72.24} \\
            \midrule
		      \multicolumn{4}{c}{\textbf{B100Inc0 (T=0)}} \\
		      \midrule
	
		      SCL-PNC (Joint Training) & \multicolumn{2}{c}{\textbf{80.44}} & \\
			\bottomrule
		\end{tabular}%
	}
	\label{Tab.4}
\end{table}

\begin{table}[h]
	\centering
	\caption{Experimental results of B50Inc10 on the ImageNet100 dataset, $t$ is the serial number of each incremental task}
	\resizebox{\textwidth}{!}{%
		\begin{tabular}{lcccccccc}
			\toprule
			\textbf{Method}  & $t$=0 &$t$=1 & $t$=2 & $t$=3 & $t$=4 &
		  $t$=5 &  $\text{Acc}_{avg}$ & PD \\
			\midrule
			LwF (PAMI’18) \cite{8107520}      & 84.40 & 47.13 & 41.60 & 38.48 & 34.42 & 31.42 & 46.24 & 52.98 \\
			iCaRL (CVPR’17) \cite{rebuffi2017icarl}      & 84.40 & 64.57 & 59.94 & 58.22 & 54.89 & 53.68 & 62.62  & 30.72 \\
			WA (CVPR’20) \cite{zhao2020fair}          & 84.40 & 67.70 & 63.66 & 62.92 & 59.56 & 56.64 & 65.81  & 27.76\\
			DER (CVPR’21) \cite{yan2021der}        & 84.40 & 80.67 & 78.34 & 76.18 & 73.80 & 71.10 & \underline{77.42}  & \textbf{13.30}\\
			FOSTER (ECCV’22) \cite{foster2022feature}      & 84.40 & 77.53 & 66.43 & 64.97 & 63.62 & 63.12 & 70.01 & 21.28 \\ 
			MEMO (ICLR’23) \cite{zhou2022memory}     & 84.40 & 80.13 & 77.71 & 75.68 & 72.82 & 70.22 & 76.83  & 14.18\\
			BEEF (ICLR’23) \cite{wang2022beef}       & - & - & - & - & - & 70.98 & 77.27 & - \\
			DS-AL (AAAI’24) \cite{zhuang2024dsal}       & - & - & - & - & - & 68.00 & 75.20  & - \\
			FCS (CVPR’24) \cite{li2024fcs}        & 81.60 & - & - & - & - & 63.82 & 74.06  & 17.78 \\
			PRL(NIPS’24) \cite{shi2024prospective}    & 84.52 & 77.90 & 72.32 & 69.72 & 67.16 & 65.44 & 72.85  & 19.08 \\	
			TagFex(CVPR’25) \cite{zheng2025task}    & - & - & - & - & - & 75.54 & \textbf{80.64}  & - \\
			SCL-PNC                         & 84.10 & 80.22 & 77.93 & 74.73 & 73.21 & 70.61 & 76.80  & \underline{13.49}\\
		
			\bottomrule
		\end{tabular}%
	}
	\label{Tab.5}
\end{table}

In the B100Inc0 scenario, SCL-PNC (Joint Training) achieved an upper bound of $\mathbf{80.44\%}$. Our incremental result in the challenging $\text{B50Inc10}$ setting ($\text{76.80\%}$) is only about $\approx \mathbf{3.64\%}$ behind this upper bound. This small gap, particularly on the complex ImageNet-100 benchmark, serves as a strong indicator that SCL-PNC’s feature alignment mechanism is highly robust in overcoming catastrophic forgetting.

Collectively, these results demonstrate that SCL-PNC attains a superior balance between computational cost and representational capacity, offering a scalable, memory-efficient, and high-performance solution for large-scale class-incremental learning.

\subsection{Ablation Study}
\label{subsec5.5}
SCL-PNC comprises three core components, which are the expandable model backbone (EM), the adapt-layer (AL), and the parametric ETF classifier. To systematically assess each module's contribution, we establish a baseline model using EM with a fully connected layer (FC) and progressively integrate or replace the different components for the comparative analysis. Table \ref{Tab.5} presents the ablation experimental results on the CIFAR-100 dataset under the B50Inc10 strategy.

\begin{table}[htbp]
	\centering
	\caption{Ablation experiment of SCL-PNC on the CIFAR-100 dataset under the B50Inc10 Strategy}
	\resizebox{\textwidth}{!}{%
		\begin{tabular}{ccccccccccc}
			\toprule
			\textbf{EM} & \textbf{AL} & \textbf{FC} & \textbf{ETF} & $t$=0 & $t$=1 & $t$=2 & $t$=3 & $t$=4 & $t$=5 & $\text{Acc}_{avg}$  \\
			\midrule
			\checkmark &             & \checkmark &             & \textbf{79.74} & 71.65 & 67.31 & 61.93 & 59.14 & 58.65 & 66.40 \\
			\checkmark &             &            & \checkmark  & 78.96 & 66.45 & 64.17 & 57.61 & 53.43 & 50.23 & 61.81 \\
			\checkmark & \checkmark  & \checkmark &             & 75.70 & 68.17 & 65.49 & 59.69 & 57.18 & 54.88 & 63.52 \\
			\checkmark & \checkmark  &            & \checkmark  & 78.62 & \textbf{74.67} & \textbf{73.04} & \textbf{68.20} & \textbf{66.28} & \textbf{64.69} & \textbf{70.92} \\
			\bottomrule
		\end{tabular}%
	}
	\label{Tab.5}
\end{table}

The experimental results demonstrate SCL-PNC with three core components outperforms other model with the part components. When a FC is used as the final layer, the incremental expansion model reaches the satisfactory classification accuracy on the base task (the 0 task). However, the classification accuracy significantly degrades in subsequent incremental tasks. In contrast, the ETF classifier located as the final layer results or the AL incorporation between the classifier and the expandable model backbone independently decline in the classification accuracy of the model. Ultimately, the complete model combination exhibits the significant performance improvements on the incremental tasks, indicating a strong synergistic effect among the components. Therefore, each component in SCL-PNC is an indispensable part for the overall classification performance.

Table \ref{Tab.7} presents the ablation results of SCL-PNC on the ImageNet-100 dataset under the B50Inc10 strategy. Overall, the performance trends are consistent with those observed on CIFAR-100, while the differences among the modules become more distinct on this larger and more complex dataset. Among all configurations, EM + AL + ETF delivers the most reliable overall performance, achieving the highest average accuracy of $76.80\%$. Although h a baseline mod(FC-based) attains the slightly higher accuracy in the base task ($\approx 85\%$), it exhibits the markedly larger degradation in later tasks, with final-task accuracy falling below $64\%$. In contrast, the EM + AL + ETF configuration maintains $70.61\%$ on the final task and surpasses FC-based baseline model by $7$--$10\%$ on average. These results demonstrate that integrating EM with AL and ETF effectively suppresses performance decay and yields a more stable and scalable incremental learning framework.
\begin{table}[htbp] 
	\centering
	\caption{Ablation experiment of SCL-PNC on the Imagenet-100 dataset under the B50Inc10 strategy}
	\resizebox{\textwidth}{!}{%
		\begin{tabular}{ccccccccccc}
			\toprule
			\textbf{EM} & \textbf{AL} & \textbf{FC} & \textbf{ETF} & $t$=0 & $t$=1 & $t$=2 & $t$=3 & $t$=4 & $t$=5 & $\text{Acc}_{avg}$  \\
			\midrule
			\checkmark &             & \checkmark &             & \textbf{85.30} & 78.32 & 73.31 & 69.26 & 66.62 & 63.11 & 72.65 \\
			\checkmark &             &            & \checkmark  & 83.48 & 79.22 & 76.89 & 73.68 & 71.98 & 61.75 & 74.50 \\
			\checkmark & \checkmark  & \checkmark &             & 85.12 & 74.50 & 70.94 & 66.43 & 63.62 & 59.68 & 70.05 \\
			\checkmark & \checkmark  &            & \checkmark  & 84.10 & \textbf{80.22} & \textbf{77.93} & \textbf{74.73} & \textbf{73.21} & \textbf{70.61} & \textbf{76.80} \\
			\bottomrule
		\end{tabular}%
	}
	\label{Tab.7} 
\end{table}

\subsection{Hyperparameter Sensitivity Analysis}
To further evaluate the robustness of SCL-PNC under the different parameters, we performed a hyperparameter sensitivity analysis on the CIFAR-100 dataset in the B50Inc10 setting. The hyperparameter is the distillation weight ($\lambda$) coefficients in equation (\ref{total_loss}).
\begin{figure}[ht]
	\centering
	\includegraphics[width=0.8\textwidth,height=0.35\textheight]{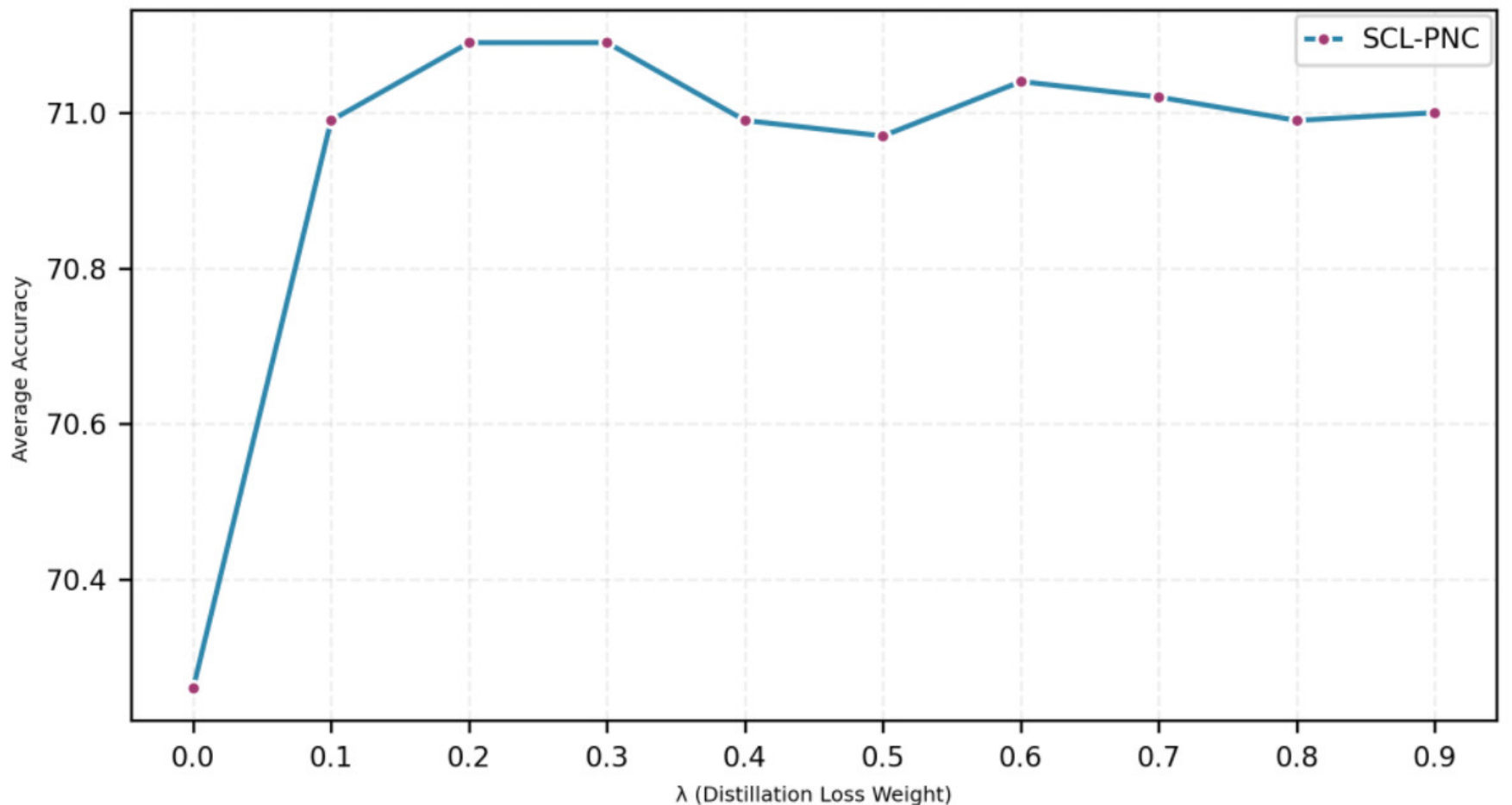}
	\caption{Hyperparameter sensitivity analysis of the CIFAR-100 dataset under the B50Inc10 strategy }\label{fig6}
\end{figure}

Figure \ref{fig6} clearly demonstrate that the proposed framework maintains high-level performance across a wide spectrum of parameter values. Crucially, the average incremental accuracy exhibits remarkable stability against hyperparameter variation form $0$ to $0.9$ with 10 intervals, which is $0.1$. Distill weights ($\lambda$) as the regularization term of the total loss, the influence of which is few by varying $\lambda$ form $0.1$ to $0.9$, and average accuracy is in the smaller observed variance, with the performance fluctuating by only $0.15\%$ percentage points ($71.09\%$ vs. $70.94\%$). When $\lambda$ is 0, the model does not involve the distill knowledge between the extend modules. This occurs the obvious decline performance (about 1\%), which shows the importance of the the distill knowledge between the extend modules.

This evidence collectively affirms the exceptional parameter robustness of the SCL-PNC framework. The model consistently yields stable, high incremental accuracy without the need for delicate and time-consuming fine-tuning of its hyperparameter settings.

\subsection{Architecture Selection of Adapt-layer}

The Adapt-layer serves as a key information processing mechanism responsible for distribution alignment between the previous and incremental classes. Consequently, the architectural selection for this layer must consider several factors: 1) compatibility with the feedforward neural network of the backbone, 2) the universal approximation capability required to fit the differences between diverse feature distributions, and 3) the
lightweight optimization of weights via backpropagation for end-to-end training.

Based on these criteria, multilayer perceptrons (MLP) and Kolmogorov-Arnold networks (KAN) \cite{somvanshi2025survey} are considered as primary candidates to model this non-linear function. To evaluate the performance influence of these different designs, we conduct the comparative experiments using the B50Inc10 strategy on CIFAR-100.
\begin{table}[ht]
    \centering
    \caption{Model performance comparison of MLP-based and KAN-based adapt-layer on B50Inc10 strategy of CIFAR100 dataset}
    \resizebox{\textwidth}{!}{%
    \begin{tabular}{lccccccc}
        \toprule
			\textbf{Method} & $t$=0 & $t$=1 & $t$=2 & $t$=3 & $t$=4 & $t$=5 & $\text{Acc}_{avg}$  \\
			\midrule
       MLP-based model & \textbf{79.12} & \textbf{74.38} & \textbf{72.54} & \textbf{67.86} & \textbf{65.73} & \textbf{63.21}  & \textbf{70.47}\\
        KAN-based model & 78.38 & 64.85 & 60.36 & 53.94 & 51.22 & 48.64 & 59.57\\
        \bottomrule
    \end{tabular}%
    }
    \label{Tab.6}
\end{table}

As shown in Table \ref{Tab.6}, the MLP-based model consistently demonstrates superior average recognition accuracy compared to the KAN-based model across all incremental stages. The main reason attributes to its better architectural compatibility with the convolutional backbone network, which provides a natural structural and functional complementarity for projecting the aligned features into the final classification space, for the superior performance of the MLP-based adapt-layer. In contrast, the KAN-based adapt-layer have the higher complexity and the slower convergence because of the flexible parameter learning of the active function to lead to the inferior performance of the the KAN-based adapt-layer.

\subsection{Similarity between Expend-layers in the Different Way}
\label{SA}
Based on the architectural hypothesis in Section \ref{subsec4.2}, which is the base-layer features serve as anchor regularizers for deep-layer representations, this section presents a comparative analysis between two backbone expansion strategies, which are the serial expansion (SE) and the parallel expansion with knowledge distillation (P-KD). In the serial expansion configuration, each newly added expand-layer sequentially connects to its predecessor, leading to accumulated information attenuation and representation inconsistency as the network depth increasing. In contrast, the parallel expansion framework allows multiple expand-layers to process information simultaneously, receiving both the general and stable features from the frozen base-layer and the adaptive representations from the previous expand-layer.

\begin{figure}[ht]
	\centering
	\includegraphics[width=1\textwidth,height=0.4\textheight]{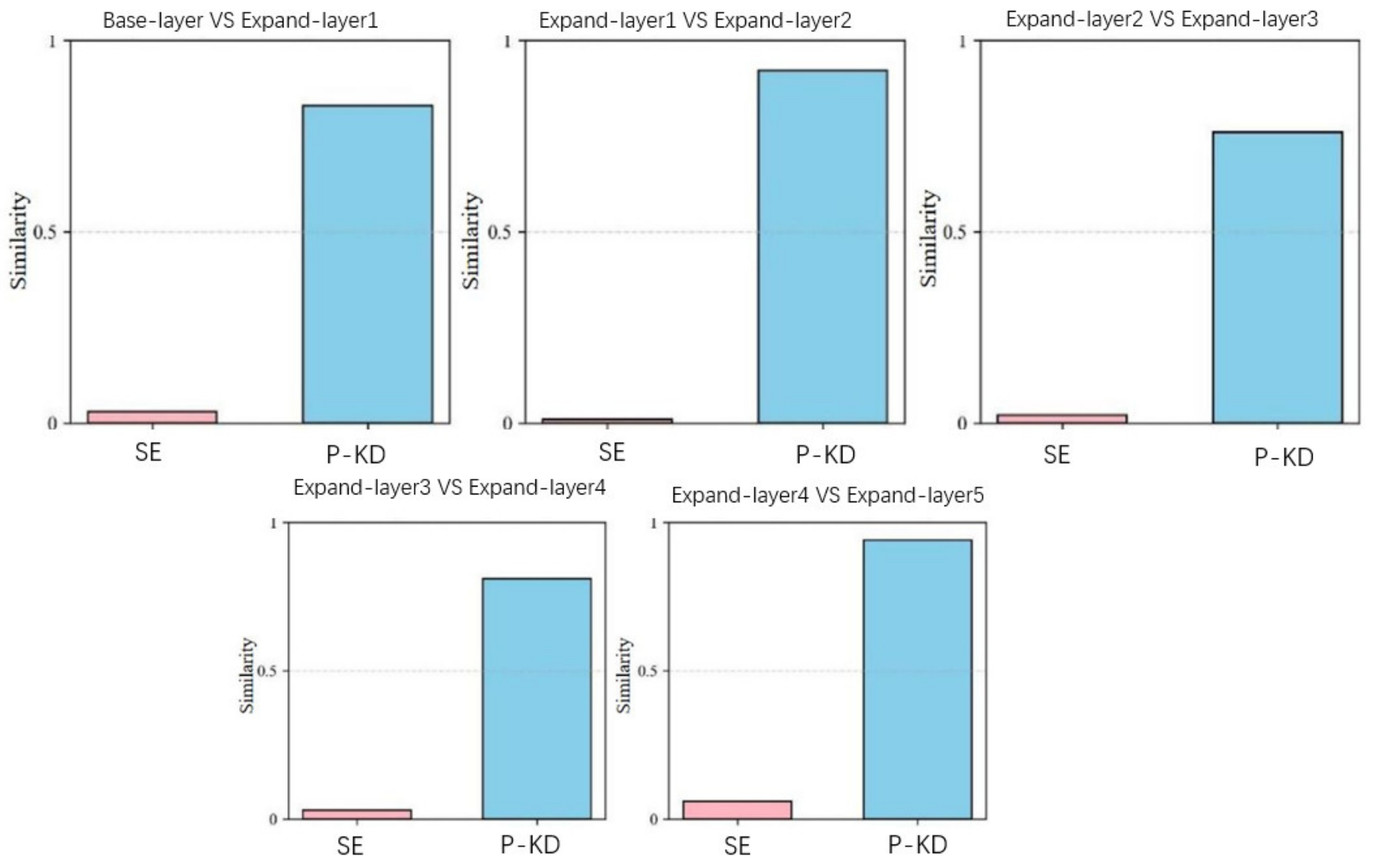}
	\caption{Similarity comparison of the features from the different expansion layers on B50Inc10 strategy of ImageNet-100 dataset. Pink indicates the feature similarity between serial expand-layers, while blue represents the feature similarity between the extend-layers of the parallel extension with knowledge distillation.}\label{fig7}
\end{figure}

To quantitatively assess the difference in feature propagation between these two configurations, we employ centered kernel alignment (CKA) [41] to measure the feature representation similarity among expansion layers. CKA serves as a reliable metric for comparing the representation geometry learned by different neural network layers, outputting a scalar value between 0 and 1, where higher values indicate greater similarity. As shown in Figure \ref{fig7}, the features from the P-KD configuration exhibit the significantly higher inter-layer similarity compared to those from SE. This striking observation confirms that the P-KD strategy effectively alleviates inter-module feature drift across expansion layers. This enhanced alignment is attributed to the parallel structure and the explicit knowledge distillation, which force the newly expanded layers to closely mimic the feature space learned by the previous, knowledge-rich layers. By ensuring consistent and aligned feature propagation, P-KD successfully mitigates the catastrophic forgetting of previously learned knowledge and promotes stable incremental learning.

\subsection{ Computational Efficiency Analysis}
\label{CEA}
To comprehensively evaluate the computational trade-offs of the SCL-PNC model in practical applications, we conduct the detailed computational efficiency analysis experiment. This experiment aims to quantify the advantages and disadvantages of SCL-PNC compared to the other expanding backbone networks methods (DER, FCS, and MEMO) in terms of resource consumption.

\begin{figure}[htbp]
    \centering
    \noindent
    \setlength{\abovecaptionskip}{8pt}
    \setlength{\belowcaptionskip}{0pt}
    
   \includegraphics[
        width=0.55\linewidth,  
        keepaspectratio,       
        trim=5 0 0 0,          
        clip                   
    ]{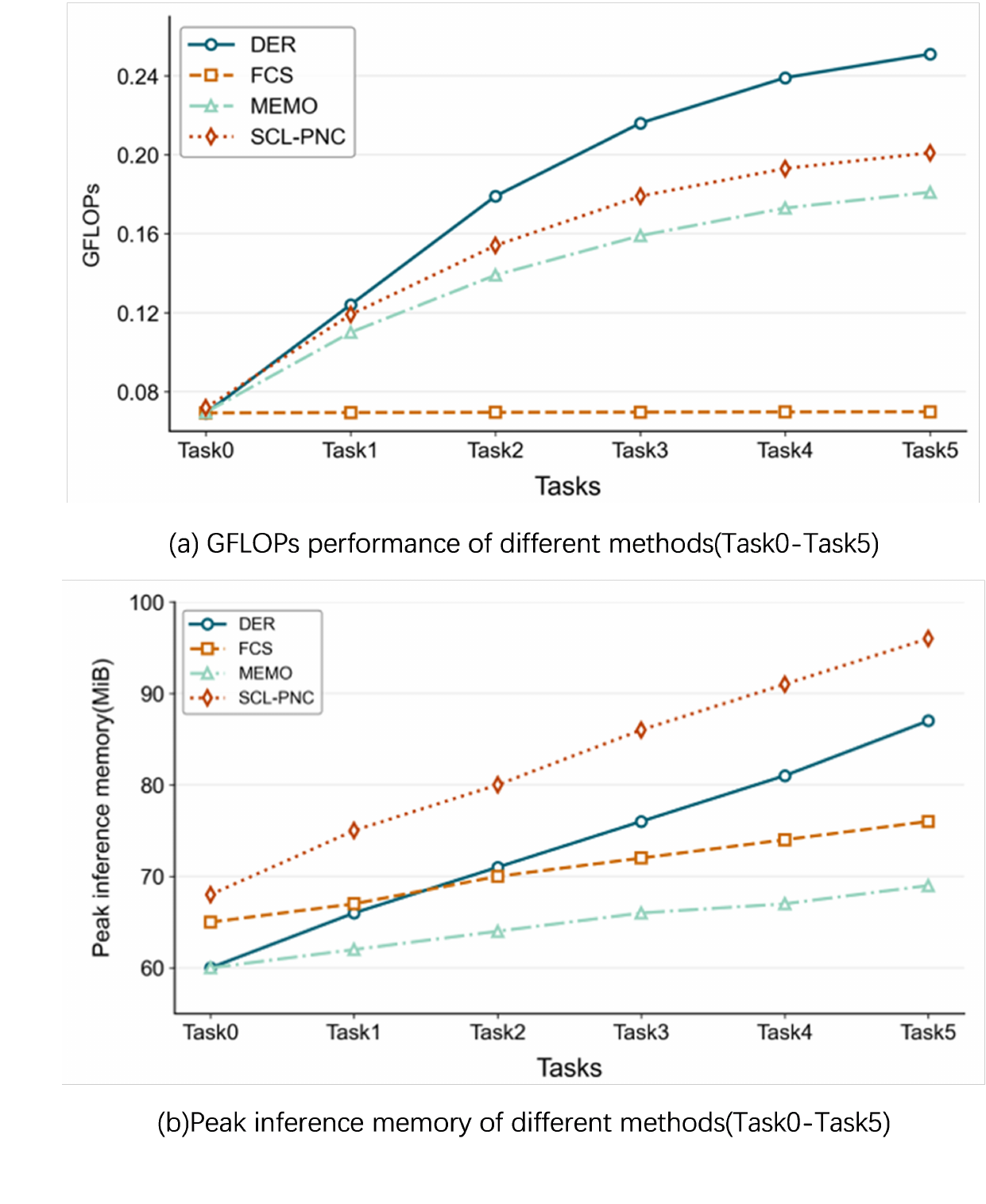}
    
    \vspace{0.00cm}  
    \includegraphics[
        width=0.55\linewidth,
        keepaspectratio,
        trim=5 0 0 0,
        clip
    ]{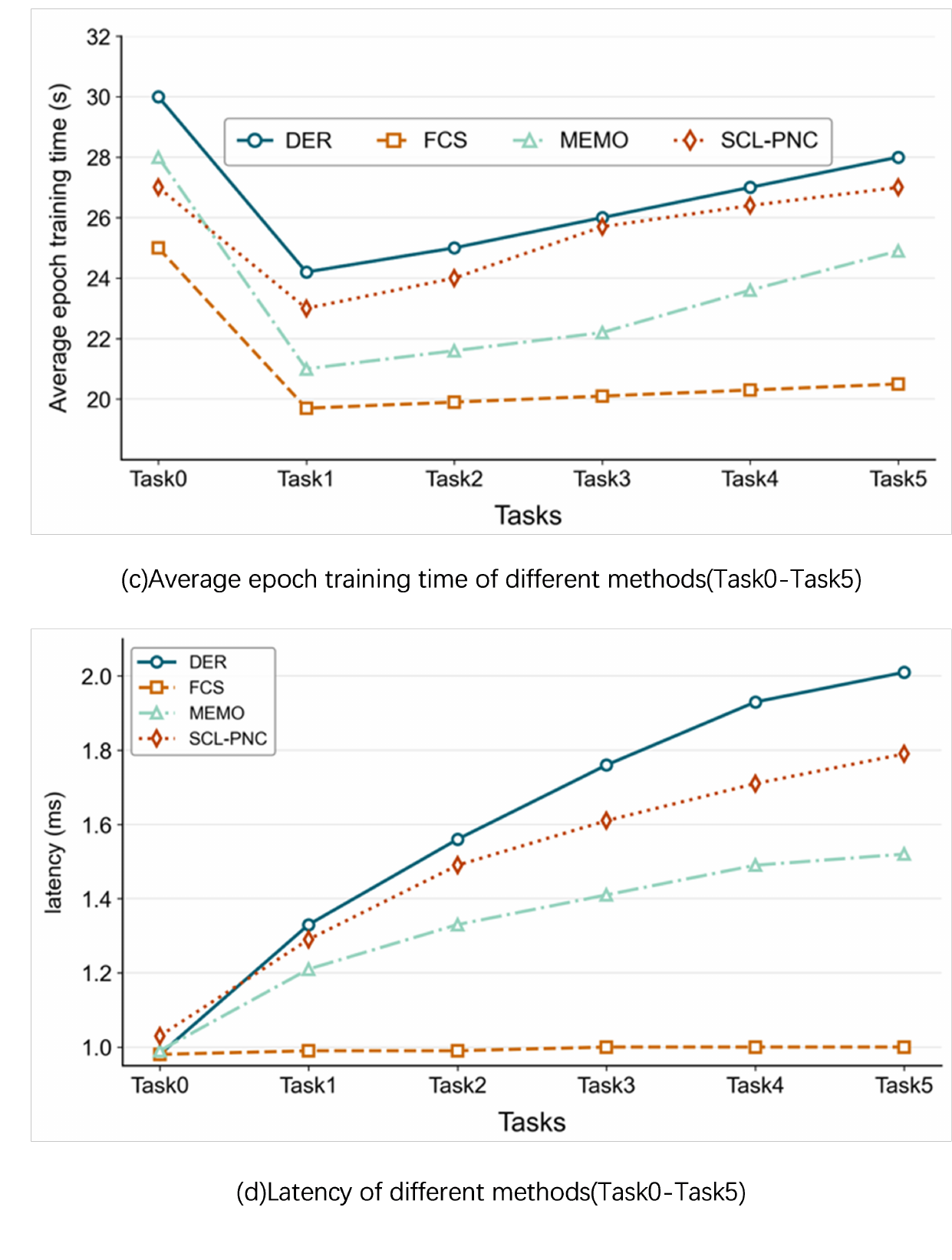}
    
    \caption{The trends of computational metrics for different incremental learning methods in the B50Inc10 scenario with the number of tasks.}
    \label{fig:fig7_2}
\end{figure}

\textbf{Experimental Setup:}
\begin{itemize}
    \item \textbf{GPU:} RTX 4090
    \item \textbf{Backbone:} ResNet32
    \item \textbf{Dataset:} CIFAR-100
    \item \textbf{Scenario:} B50Inc10 (10 tasks, 10 classes added per task)
    \item \textbf{Metrics:} We collected the model's computational metrics after completing all tasks. These metrics include giga floating-point operations per second (GFLOPs) during inference, peak inference memory during inference, average epoch training time and latency during inference.
\end{itemize}

The dynamic metrics under different tasks and their trends with the number of tasks are shown in Figure \ref{fig:fig7_2}. As illustrated in Figure \ref{fig:fig7_2}(a) and (d), the GFLOPs and latency of SCL-PNC and DER demonstrate a clear linear scaling with the increasing number of tasks. This observation explicitly confirms that the use of parameter expansion mechanisms (such as the expansion layer in SCL-PNC) inevitably leads to an increase in the model's inference computational complexity as the task count growing.

Despite this linear scaling, the absolute values of GFLOPs and latency for SCL-PNC consistently remain lower than those for DER throughout the incremental learning process. This disparity strongly suggests that SCL-PNC achieves superior efficiency in maintaining model compactness compared to DER.

Conversely, the curves representing FCS and MEMO show near-constant, flat performance. This confirms that these methods do not introduce additional computational paths and possess an absolute advantage in inference speed. However, this inferential speed advantage these methods does not translate into the competitive accuracy and the adaptability in the different scenarios when benchmarked against the SCL-PNC method.

\section{Conclusion}
\label{sec6}

Recent advances in incremental learning algorithms have gained significant momentum, particularly with the advent of large-scale models and datasets, attracting the growing attention from the research community. We proposed the SCL-PNC, a neural collapse-induced expandable model for class incremental Learning, which demonstrates the superior classification performance compared to state-of-the-art methods. To effectively manage feature drift in expandable architectures, we introduced a novel knowledge distillation mechanism between successive expand-layers and integrated a parametric ETF classifier with an adapt-layer for robust cross-task class distribution alignment. This comprehensive strategy effectively mitigates feature drift and substantially improves the retention of previously acquired knowledge.

Nevertheless, while SCL-PNC demonstrates superior overall accuracy and scalability, its reliance on sequential expansion and the associated knowledge preservation strategy introduces several new challenges for long-term efficiency and adaptability. Specifically, our current approach is limited by the following factors. The framework introduces an independent Expand-layer for each incremental task. Although each layer is lightweight, the total parameter count scales linearly with the number of tasks $(T)$, leading to significant storage pressure in very long-term learning sequences. Knowledge-distillation mechanism, while effective, introduces accumulating computational cost and latency across extended incremental sequences. The accumulation of inter-task distribution shifts may gradually weaken the Adapt-layer's ability to generalize effectively across all learned categories, potentially limiting feature plasticity. The model's generalization ability depends on the diversity of classes in the base task. When the base task is small or biased, the learned representations may not be sufficiently generic, causing reduced performance in later stages compared with parameter-reuse approaches.

Future works will be guided by the observation that selective parameter retention, rather than complete parameter freezing, may unlock better performance stability. This suggests that allowing selective parameter updates could be a highly promising research direction. To mitigate the linear parameter scaling, we will explore parameter sharing mechanisms to decouple the total model size from the number of incremental tasks.To reduce the accumulative computational overhead, we will investigate lightweight distillation schemes, such as selective or knowledge-efficient distillation. Furthermore, we will focus on designing more robust Adapt-layers to enhance their cross-task generalization ability against severe distribution shifts. Additionally, we will continue to develop dynamic parameter selection mechanisms that intelligently balance stability and plasticity, potentially through attention-based or gradient-sensitive approaches.

\section*{Declaration of competing interest}
The authors declare that they have no known competing financial interests or personal relationships that could have appeared to
influence the work reported in this paper.
\section*{Acknowledgments}
 This work was supported by the National Natural Science Foundation of China (NSFC, Program No. 61771386), Key Research and Development Program of Shaanxi (Program No. 2020SF-359) and Natural Science Basic Research Plan in Shanxi Province of China (Program No. 2021JM-340).

\end{document}